\pdfoutput=1

\documentclass[11pt]{article}

\usepackage[]{acl}

\usepackage{eso-pic}
\newcommand{\citestamp}{
  \AddToShipoutPicture*{%
    \setlength{\unitlength}{1mm}
    \put(108,14){\makebox(0,0){\footnotesize {\em Proceedings of the
        62nd Annual Meeting of the Association for Computational Linguistics (Volume 1: Long Papers)}, pages 13806--13834}}
    \put(108,10){\makebox(0,0){\footnotesize August 11--16, 2024. \textbf{Updated version of 17 June 2024.} \copyright 2024 Association for Computational Linguistics.}}
  }
}

\usepackage[group-separator={,},separate-uncertainty=true]{siunitx}
\usepackage{times}
\usepackage{latexsym}
\usepackage{arydshln}
\usepackage{fontawesome}
\usepackage{graphicx,subcaption}
\usepackage{amsmath,amssymb,amsfonts}
\usepackage{xspace}
\usepackage{color}
\usepackage{enumitem}
\usepackage{comment}
\usepackage{mathabx}
\usepackage[T1]{fontenc}

\usepackage[utf8]{inputenc}

\usepackage{microtype}

\usepackage{inconsolata}
\usepackage{amsmath}
\usepackage[disable]{todonotes}
\makeatletter
\newcommand*\iftodonotes{\if@todonotes@disabled\expandafter\@secondoftwo\else\expandafter\@firstoftwo\fi}
\makeatother
\newcommand{\noindentaftertodo}{\iftodonotes{\noindent}{}\ignorespaces}

\newcommand{\Fixme}[2][]{\noindentaftertodo}
\newcommand{\Notewho}[3][]{\noindentaftertodo}
\newcommand{\Helia}[2][]{\noindentaftertodo}
\newcommand{\Chris}[2][]{\noindentaftertodo}
\newcommand{\Jason}[2][]{\noindentaftertodo}
\newcommand{\Ben}[2][]{\noindentaftertodo}

\usepackage{multirow} 
\usepackage{xspace}

\usepackage[noabbrev,capitalize]{cleveref}  

\crefname{page}{page}{pages}
\crefname{footnote}{footnote}{footnotes} 
\crefname{equation}{equation}{equations} 
\crefname{section}{\S}{\S\S}
\Crefname{section}{\S}{\S\S}
\crefformat{section}{#2\S#1#3}
\Crefformat{section}{#2\S#1#3}
\crefrangeformat{section}{\S\S#3#1#4--#5#2#6}
\Crefrangeformat{section}{\S\S#3#1#4--#5#2#6}
\crefmultiformat{section}{\S#2#1#3}{ and~\S#2#1#3}{, \S#2#1#3}{ and~\S#2#1#3}
\Crefmultiformat{section}{\S#2#1#3}{ and~\S#2#1#3}{, \S#2#1#3}{ and~\S#2#1#3}
\crefrangemultiformat{section}{\S\S#3#1#4--#5#2#6}{ and~\S\S#3#1#4--#5#2#6}{, \S\S#3#1#4--#5#2#6}{ and~\S\S#3#1#4--#5#2#6}
\Crefrangemultiformat{section}{\S\S#3#1#4--#5#2#6}{ and~\S\S#3#1#4--#5#2#6}{, \S\S#3#1#4--#5#2#6}{ and~\S\S#3#1#4--#5#2#6}

\newcommand{\llmeval}{\textsc{LLM-Rubric}\xspace}
\newcommand\numCriteria[0]{8}
\newcommand\numCriteriaAndOverall[0]{9}
\newcommand{\defeq}{\mathrel{\stackrel{\textnormal{\tiny def}}{=}}}
\newcommand{\Y}{\mathcal{Y}}
\newcommand{\T}{\mathcal{T}}
\newcommand{\A}{\mathcal{A}}
\newcommand{\vx}{\mathbf{x}}
\newcommand{\vz}{\mathbf{z}}
\newcommand{\vw}{\mathbf{w}}
\newcommand{\yhat}{\hat{y}}
\newcommand{\phat}{\hat{p}}
\newcommand{\rhat}{\hat{r}}
\newcommand{\pllm}{p_{\mathrm{LLM}}}
\newcommand{\dataset}{\mathcal{D}}
\DeclareMathOperator*{\mean}{mean}
\DeclareMathOperator*{\argmax}{argmax}

\newcommand{\horizbar}{\medskip\noindent\rule{\textwidth}{1pt}\medskip\par\noindent}

\title{\llmeval: A Multidimensional, Calibrated Approach \\ to Automated Evaluation of Natural Language Texts$^\dagger$
}

\author{Helia Hashemi$^*$\quad$\!$
Jason Eisner$^*$\quad$\!$
Corby Rosset\quad$\!$
Benjamin Van Durme\quad$\!$
Chris Kedzie
\\
Microsoft  \\
\texttt{\{hhashemi,jeisner,corbyrosset,ben.vandurme,chriskedzie\}@microsoft.com}
}

\begin{document}
\maketitle%
\def\thefootnote{*}\footnotetext{Equal contribution.}\def\thefootnote{\arabic{footnote}}%
\def\thefootnote{$\dagger$}\footnotetext{Code and data available at \url{https://github.com/microsoft/llm-rubric}.}%
\def\thefootnote{\arabic{footnote}}

\begin{abstract}
This paper introduces a framework for the automated evaluation of natural language texts. A manually constructed rubric describes how to assess multiple dimensions of interest.  To evaluate a text, a large language model (LLM) is prompted with each rubric question and produces a distribution over potential responses.  The LLM predictions often fail to agree well with human judges---indeed, the humans do not fully agree with one another. However, the multiple LLM distributions can be \emph{combined} to \emph{predict} each human judge's annotations on all questions, including a summary question that assesses overall quality or relevance. \llmeval accomplishes this by training a small feed-forward neural network that includes both judge-specific and judge-independent parameters. When evaluating dialogue systems in a human-AI information-seeking task, we find that \llmeval with $\numCriteriaAndOverall$ questions
(assessing dimensions such as naturalness, conciseness, and citation quality) predicts human judges' assessment of overall user satisfaction, on a scale of 1--4, with RMS error $< \num{0.5}$, a $2\times$ improvement over the uncalibrated baseline.
\looseness=-1
\end{abstract}

\citestamp

\section{Introduction}
\label{sec:intro}

Many fields that must assess large numbers of short documents have turned to NLP-assisted workflows.  For example, 
lawyers conducting legal discovery must identify all relevant documents \cite{TAR-2019}---a task also faced by journalists and historians.  
Social scientists and market researchers must code survey responses (\citealp{mellon2024ais}; \href{https://www.enumerate.ai/}{enumerate.ai}; \href{https://atlasti.com/}{ATLAS.ti}).
Teachers or examiners must evaluate student writing \cite{page-1968,ramesh-sanampudi-2022} and provide feedback \cite{meyer2024feedback}.  
Doctors, social workers, or public health agencies or researchers may assess an individual's mental health or safety from their social media posts \cite{chancellor2020methods,xu2024mentalllm,algaradi2022natural} or from clinical interviews and assessments \cite{galatzerlevy2023capability}.

The above settings evaluate human-authored texts.  In addition, NLP developers must assess the quality of their machine-generated texts---texts that are consumed by end users, but also hidden intermediate steps in agentic workflows (such as chains of thought, tool calls, and revisions).  With the recent commercialization of conversational AI, for example, it is crucial to evaluate dialogue systems during development and monitor them after deployment.  Special care is needed in high-stakes settings like medical dialogue \cite{huang2024assessment}.  

Manual evaluation has long been the gold standard for assessing text, including generated text \citep{Saphra2023FirstTT,human_eval:2021}.
Humans are often asked to consider multiple criteria and then provide a final assessment \cite{Hosking2023HumanFI}.
Humans may also be asked to produce reference answers to which other humans can compare the target text.  Yet manual evaluation is expensive, time-consuming, and not without its own quality and reliability issues \citep{Hosking2023HumanFI,liu-etal-2016-evaluate,smith-etal-2022-human}.  Because of these challenges, and the increasing abilities of large language models (LLMs) \cite{NEURIPS2020_1457c0d6}, experimenters have recently 
been eliciting ratings directly from an LLM (\citealp{chiang-lee-2023-large,fu2023gptscore,liu-etal-2023-g,thomas2024large}; \href{https://chainforge.ai/docs/evaluation/#llm-scorers}{ChainForge}; and others). \emph{But can LLM evaluation be trusted?} It solves the time, scaling, and possibly cost issues, but leaves open the problem of aligning these LLM ratings with human judgments.

\begin{figure*}
    \centering
    \includegraphics[width=\textwidth,clip,trim="0cm 6.5cm 0cm 0cm"]{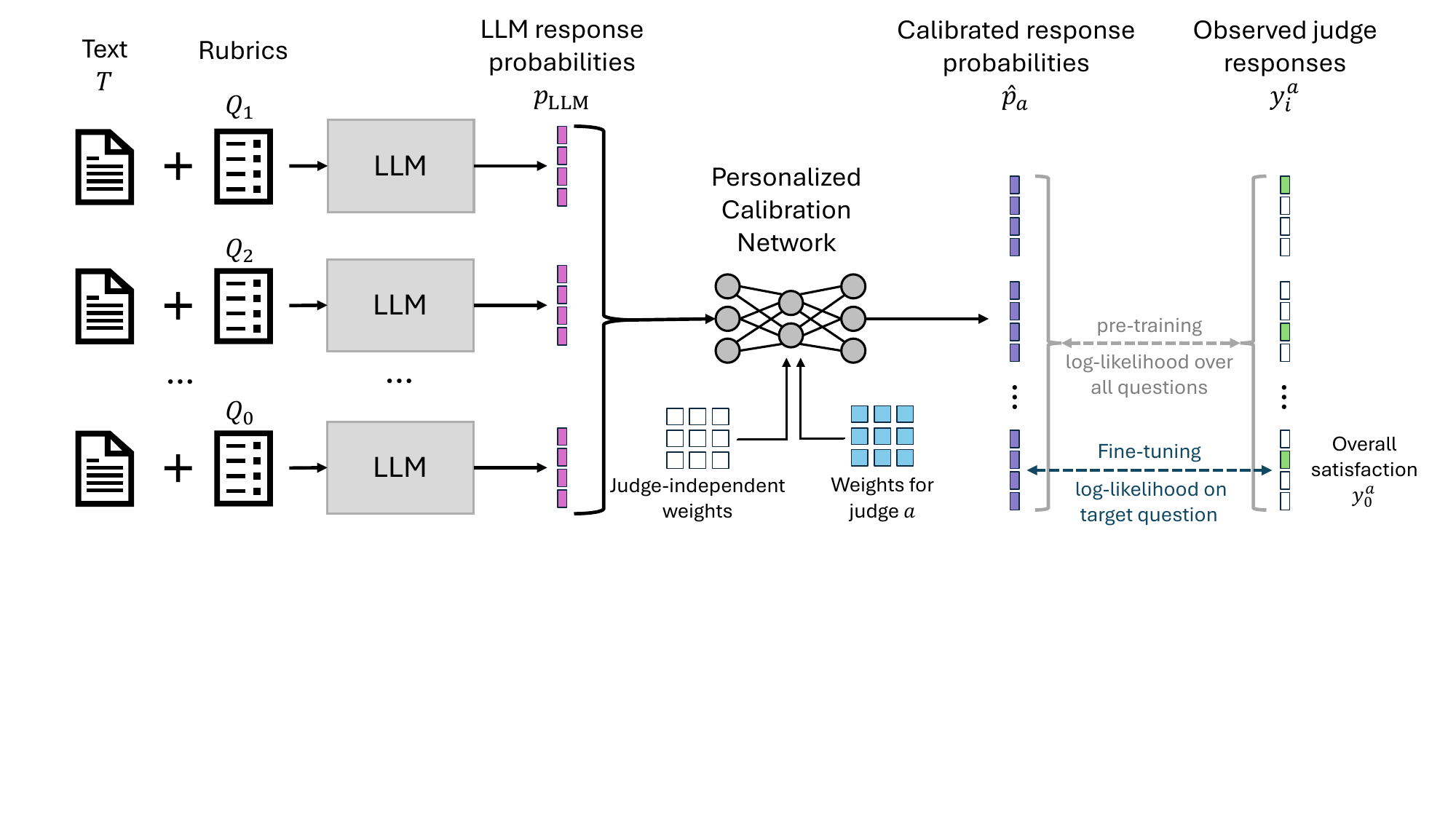}
    \caption{An overview of the \llmeval framework. The LLM and its prompts are fixed across texts and judges, but the calibration network weights are trained to predict the responses of various human judges.}\label{fig:llm_rubric}
    \vspace{-4mm}
\end{figure*}

We present a general approach to this alignment problem.  We demonstrate its value for the evaluation and comparison of LLM-powered dialogue systems, in an information-seeking dialogue task \citep{info-seeking-conv} similar to \citet{lowe-etal-2015-ubuntu}.
Evaluation in this setting is complex owing to competing factors that might affect a human judge's assessment of the dialogue. These may include correctness of responses, accuracy and helpfulness of citations, length and complexity of responses, and more  \citep{smith-etal-2022-human}.  

Our \llmeval approach begins with a manually authored evaluation rubric.  The rubric's multiple-choice questions cover various evaluation dimensions, and it may also include a question that assesses \emph{overall} quality or relevance.
Evaluating a text, such as a dialogue, then consists of two main steps: (1)~for each rubric question we elicit the LLM's probability distribution over possible responses, by prompting it with the text and the rubric question, and (2)~we aggregate and calibrate these distributions with a small feed-forward network that has been trained to match the individual preferences of human judges. A high-level overview of \llmeval is shown in \cref{fig:llm_rubric}.

For research in generative NLP, once the rubric and LLM are fixed, \llmeval can be used like other metrics (\textsc{Bleu}, \textsc{Rouge}, etc.) to drive system development, monitor quality, demonstrate the value of a new technique, and conduct competitions.  In our dialogue evaluation experiments, each user--AI dialogue is evaluated by 3 trained annotators (randomly drawn from a larger pool) who each answered the same 9 rubric questions.  Our method uses these data to train an automatic LLM-based evaluator, without treating the 24 human annotators as interchangeable.
Overall, we find\footnote{See \cref{tab:main_results}, right side, rows 3, 4, and 6.} that 
\begin{itemize}[noitemsep]
\item Personalized calibration of an LLM evaluator of overall satisfaction on $<$ \num{750} synthetic dialogues significantly improves its prediction of human judgments and correlations with human judgments, but still works poorly.
\item Incorporating LLM evaluations of \num{8} additional criteria (\llmeval) improves these metrics by over $2\times$ over the uncalibrated LLM.
\end{itemize}

Accurate automated text assessment could replace human assessment in many other settings, such as those reviewed at the start of this paper.  It could also be used in new settings where human assessment was never feasible.  In AI-powered user interfaces, instantaneous scoring of user-written text can feed into downstream decisions such as providing writing feedback or deciding how to proceed with a dialogue.  
An AI reasoning engine may internally apply a rubric to assess the validity of a proposed natural-language reasoning step \cite{weir2024enhancing}.
When processing a large document collection, an LLM can be used to assess the compatibility of two text passages \cite{zhang-etal-2023-clusterllm,viswanathan2023large,choi2024factgpt}, potentially in a more nuanced way than vector similarity; this problem arises in workflows for matching, routing, clustering, and fact-checking (\citealp{Charlin2013TPMS,harman-1996}; and the papers just mentioned).  
Finally, automated assessments could provide signals for \emph{training} text generation \cite{keskar-et-al-2019,tambwekar2019controllable,bai2022constitutional}.

To allow \llmeval to support all of these use cases, we release general code along with the datasets we created for this paper (see URL on page 1).  We discuss limitations at the end of the paper.

\begin{figure*}
    \centering
     \includegraphics[width=\textwidth,clip,trim="28pt 0pt 60pt 0pt"]{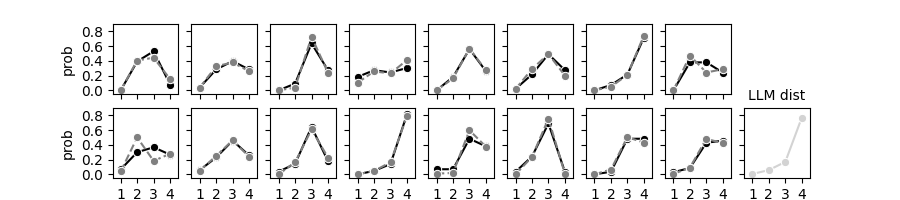}
    \caption{Our calibration network learns how different human judges use the response range 1--4.  Each black curve shows a different judge's distribution of responses to the ``overall satisfaction'' question $Q_0$ on our synthetic conversation dataset.  (We show the judges who evaluated $\geq 30$ conversations.) The corresponding gray curve shows the average distribution predicted for that judge on the same dialogues by 
    \llmeval (using cross-validation).  The final curve in light gray shows the original uncalibrated distribution of responses to $Q_0$ by the LLM (\texttt{gpt-3.5-turbo-16k}).} 
  \label{fig:judge_scores_dist}
\vspace{-.5cm}
\end{figure*}

\section{The \llmeval Method}
\label{sec:method}

It is challenging to model human preferences in a combinatorial space such as text.  Reasonable human judges may differ \cite{aroyo-welty-2015} on (1) what textual properties they happen to prefer (e.g., concise vs.\@ detailed, formal vs.\@ informal, novice vs.\@ expert audience), (2) how they combine multiple preferences into an overall assessment, and (3) how they convey that assessment through a numerical score.  \Cref{fig:judge_scores_dist} shows that in our dataset (\cref{sec:data}), different human judges indeed have very different marginal distributions of overall score.  Clearly these cannot all be matched by a judge-independent system (e.g., the LLM shown at the lower right of \cref{fig:judge_scores_dist}).

To expose the different properties and preferences at play, we ask the human judges a series of finer-grained questions about different evaluation criteria.  It is already common in practical settings (\cref{sec:intro}) to at least mention such criteria in instructions to human judges.  We use the same questions to query an LLM,\footnote{\label{fn:differentq}It is convenient to use the \emph{same} questions, as we have already crafted them. However, different or additional questions could in principle be used---or multiple variants of each question, or multiple LLMs.  This could potentially provide more useful evidence to the calibration network below, at the cost of slowing down evaluation and at the risk of overfitting.}
and train a calibration network to jointly adjust the LLM's scores to match the scores of any given human judge.  We refer to this methodology as \llmeval.  The gray curves in \cref{fig:judge_scores_dist} show that on held-out dialogues, the calibrated overall score is now distributed like that of the given judge.  We will see later that these scores are also more accurate on the individual dialogues.

In this section, 
we present \llmeval in a general way, but for concreteness, we also introduce details of our specific experimental setup.

\paragraph{Evaluation Rubric Construction.} 

We wrote $\numCriteria$ dialogue evaluation questions ($Q_1,\ldots, Q_\numCriteria$)
inspired by
the NLG evaluation literature \cite{zhou-etal-2022-deconstructing,human_eval:2021}. These questions are shown in \cref{app:questions}.  They 
 address various dimensions such as naturalness, relevance, attribution, citation quality, and conciseness.
Our final question ($Q_0$) asked the judge to assess the overall quality of the dialogue (in this case, focusing only on whether the user would be satisfied), on a Likert scale of 1--4.
Each question stated its allowed multiple-choice responses (usually scores 1--4, with a meaning provided for each score).  

\paragraph{Multi-Dimensional Evaluation with LLMs.}
We use an LLM to evaluate a given text $T$ (in our case, a dialogue transcript).
For each question $Q_i$ ($0 \leq i \leq \numCriteria$ in our case), 
we instruct the LLM to generate a label $y_i \in \Y_i$, where $\Y_i$ is the
set of allowed responses to $Q_i$ (e.g., $\{\texttt{``1''},
\texttt{``2''}, \texttt{``3''}, \texttt{``4''}\}$).
Specifically, we prompt it with a preamble, the text $T$, and the question $Q_i$, where $Q_i$ also specifies the allowed responses $\Y_i$ (see \cref{app:prompt:llm}).
We chose to do this independently for each question $Q_i$ to avoid confounding the LLM's responses.  We thus obtain $\pllm(y_i \mid T, Q_i)$ for all questions $Q_0,\dots,Q_\numCriteria$ and each possible response
$y_i \in \Y_i$.\footnote{\label{fn:unnorm}The LLM also allocates some probability to responses outside $\Y_i$, so $Z_i \defeq \sum_{y_i \in \Y_i} \pllm(y_i \mid T, Q_i) < 1$.  We do not normalize the probabilities by $Z_i$ before presenting them to the calibration network. This allows our calibration network, in principle, to notice when $Z_i \ll 1$ and to learn not to rely on the LLM's answer to $Q_i$ in such cases.  In practice, however, our prompts result in $Z_i$ being very close to 1.}

\paragraph{Aggregated Evaluation with Personalized Calibration.}
We then use a small feed-forward  \emph{calibration network} (\cref{fig:llm_rubric} and \crefrange{eq:calibfirst}{eq:softmax} below) to map this collection of LLM probabilities $\pllm(y_i \mid T, Q_i)$ to a collection of adjusted probabilities $\phat_a(y_i \mid T, Q_i)$ that predict the responses of a particular judge $a$.  Note that each $\phat_a(y_i \mid T, Q_i)$ is predicted from the LLM's behavior on \emph{all questions} about $T$, not just $Q_i$.  This design lets the calibration network inspect some additional properties of $T$ that might influence $a$'s response to $Q_i$.\footnote{\label{fn:embedding}In the future, for this reason, the calibration network's input could also include an embedding of the full text $T$.} This design also 
extends to the case where the LLM was not asked the specific question $Q_i$ for which we are predicting $a$'s response (see \cref{fn:differentq}).  

We train the calibration network by maximum likelihood (regularized by early stopping). That is, given a dataset $\dataset$ of annotations, we maximize\footnote{\label{fn:nonindep}This formula models the $y_i^a$ for different $i$ as conditionally independent given $T$.  This assumption could be relaxed.  For example, perhaps all of the $y_i^a$ should be made to also depend on a latent variable, e.g., judge $a$'s mood while annotating $T$.} 
\begin{equation}\label{eq:loglik}
\sum_{(T,i,a,y_i^a) \in \dataset} \log \phat_a(y_i^a \mid T, Q_i)
\end{equation}
where $(T,i,a,y_i^a) \in \dataset$ means that judge $a$ answered $Q_i$ on $T$ with response $y_i^a$.  

\paragraph{Decoding.}
Given a new text $T$, the trained calibration network predicts any judge $a$'s possible responses to question $Q_i$ via the \emph{distribution} $\phat_a(y_i \mid T, Q_i)$.  If we wish to output a \emph{single} predicted value $\yhat_i^a$ for downstream use, then we also need a \emph{decoding principle} that extracts $\yhat_i^a$ from $\phat_a$.  In our experiments, actual responses $y_i^a$ are integers, predictions $\yhat_i^a$ are real numbers, and we will be evaluating the predictions by $L_2$ loss, $(\yhat_i^a - y_i^a)^2$.\footnote{\label{fn:interval}This setup treats the integers as falling on an interval scale, not just an ordinal scale.  For example, outputting 1.4 when the true answer is 1 is considered exactly as bad as outputting 2.6 when the true answer is 3.  This is not always appropriate.}  Thus, our principle is to minimize the \emph{expected} $L_2$ loss (our ``Bayes risk'').  This is accomplished simply by predicting the mean of distribution $\phat_a$,
\begin{align}\label{eq:mbr}
\yhat_i^a &= 
\sum_{y_i \in \Y_i} \phat_a(y_i \mid T, Q_i) \cdot y_i
\end{align}

We remark that we could have constructed a network that directly predicted the $\yhat_i^a$ values, and trained it to minimize $L_2$ loss on training data---a regression problem.  However, by modeling the entire distribution $\phat_a$ and not just its mean, we make fuller use of the training data for representation learning---our representations are trained to be able to predict the full distribution.  Indeed, we found in pilot experiments that our method slightly outperforms the regression method.  Furthermore, modeling $\phat_a$ lets us report our predictive uncertainty---e.g., the entropy or variance of $\phat_a(y_i \mid T,Q_i)$ and not just its expectation $\yhat_i^a$.  Finally, \cref{eq:mbr} nicely guarantees that $1 \leq \yhat_i^a \leq 4$ on any example.

\paragraph{Calibration Network Architecture.}
Our network's input is a feature vector 
$\vx = \left[\pllm(y_i\mid T, Q_i) : i \in \{0,\ldots,\numCriteria\}, y_i \in \Y_i \right]$.  These are already \emph{extremely} high-level text features, extracted by the LLM.
We next use a feed-forward neural net to transform $\vx$ into a representation $\vz_2 \in \mathbb{R}^{h_2}$:
\begin{align}
     \vz_1 &= \sigma\big(\left(W_1 + W_1^a\right) [1; \vx]\big) \in \mathbb{R}^{h_1} \label{eq:calibfirst} \\
     \vz_2 &= \sigma\big(\left(W_2 + W_2^a\right) [1; \vz_1]\big) \in \mathbb{R}^{h_2} 
\end{align}
Here $W_1,W_1^a \in \mathbb{R}^{h_1 \times (1+\numCriteriaAndOverall)}$ and $W_2,W_2^a \in \mathbb{R}^{h_2 \times (1+h_1)}$.  The parameters $W_k$ are shared across all judges while $W_k^a$ are judge-specific.

The learned representations $\vz_2$ are shared across all questions.  For each $i \in \{0,\ldots,8\}$, we obtain $\{\phat_a(y_i \mid T, Q_i) : y_i \in \Y_i\}$ as a probability vector
\begin{equation}\label{eq:softmax}
\mathrm{softmax}(\left(V_i + V_i^a\right)[1; \vz_2]) \in \mathbb{R}^{|\Y_i|} \end{equation}
The collection of matrices $V_i \in \mathbb{R}^{|\Y_i| \times (1+h_2)}$ can be implemented as a 3D tensor $V$ (padding $V_i$ with extra rows when $|\Y_i|$ is small).

\paragraph{Multi-Task Learning.}  Our calibration network performs multi-task learning: each rubric question is a different task.  When the accurate prediction of $y^a_0$ is our \emph{main task}, the other tasks serve only as regularizing \emph{auxiliary tasks}, which help training to discover useful hidden features $\vz_2$.  The weighting of the auxiliary tasks could be dynamically adapted during training (using a validation set), for example with the AuxiNash training algorithm \cite{shamsian2023auxiliary}. 
However, we currently use a simpler, faster shortcut that divides training into two phases.
In the \textbf{pre-training} phase, we optimize the full log-likelihood objective \labelcref{eq:loglik}.  This learns useful initial representations.\footnote{However, in contrast to AuxiNash, this shortcut does not try to identify and favor more useful auxiliary tasks.  \Cref{eq:loglik} simply weights each question $Q_i$ in proportion to its number of annotated answers in the training dataset $\dataset$.  (In our experiments, all questions are equally represented in $\dataset$.)}  In the \textbf{fine-tuning} phase, we continue training with a modified objective that sums over only the tuples in $\dataset$ with $i=0$.  This adjusts the parameters to focus on the main task---predicting responses $y_0^a$ to $Q_0$.  In both phases, we use early stopping to avoid overfitting.\footnote{We also tried a variant where pre-training was itself divided into two stages and we fixed  $W_k^a=0$ and $V_i^a=0$ during the first stage.  This was intended to prevent overfitting of these judge-specific parameters, but we observed no improvement compared to the simpler method.}

\paragraph{Using the Predictions.} Since \llmeval can predict any judge's scores on a new text $T$, how should it be used in practice?  In \cref{app:agg}, we propose approaches to score aggregation, system quality monitoring, and other practical issues.

\paragraph{Future Extensions.}
The idea of a calibration network is quite general and can easily be extended to various types of human and LLM ratings. In \cref{app:extensions}, we sketch some natural extensions that were not needed in this paper’s experiments.

\section{Data}
\label{sec:data}

Conversational AI systems are now being widely deployed.  To test our methods on dialogue evaluation, we invest in developing both synthetic and real datasets of human--AI conversations. 

We focus on 
English information-seeking dialogues in the ``IT help'' (enterprise information technology) domain \cite{lowe-etal-2015-ubuntu,10.1007/11677482_3}.
As in many real world domains, dialogue data here is often proprietary to the system owner and/or private to the user.
Acquiring experimental access to live systems for evaluation is even more difficult. Thus, we build and evaluate several LLM-powered dialogue systems, which differ in their ability to search a corpus of websites related
to Microsoft Azure\footnote{\url{https://azure.microsoft.com/}} help topics.

For \emph{training} data, we generate a corpus of \emph{synthetic} dialogues with simulated users, and have human judges rate them. Collecting these artificial dialogues is efficient, since judges only have to annotate conversations and not interact with the systems first. 
For our final \emph{test} data, we have our judges actually interact with the live systems as users and then annotate their own dialogues.
All of our judges are professional annotators.

To mine the topics for both synthetic and live evaluation, we use real user queries and click data from a large commercial web search engine,
which further increases the realism of our experiments.

Below, \cref{sec:data:log_mining} explains how we compile a corpus of background documents and how we select topics to ensure that the generated and collected conversations are diverse and are indeed information-seeking, rather than navigational or transactional. \Cref{sec:data:synthetic,sec:data:real} explain our approaches to synthetic dialogue generation and real dialogue collection.

\subsection{Mining Topics for RAG}
\label{sec:data:log_mining}
To simulate or collect diverse information-seeking dialogues, we need to know
what information our users will seek.
We picked an arbitrary IT help topic, Azure, for which many answers can be found on the subreddit \href{https://www.reddit.com/r/AZURE/}{r/azure}. We hypothesize that search queries are enterprise information-seeking topics related to Azure if they lead to satisfactory clicks on the Azure subreddit.\footnote{A satisfactory click in a search engine is defined as a click that leads to a dwell time longer than a given threshold \citep{jiang-2016-reducing}. Here we use a threshold of \num{30} seconds.} Using this heuristic to help filter query logs obtained from the Bing search engine, we construct a set $\mathcal{S}$ of \num{2275} common English queries about Azure.  We will use these as \emph{topics} to prompt the creation of 
realistic and diverse conversations.

Some of our dialogue systems will condition their responses on relevant documents, as in retrieval-augmented generation (RAG) \cite{lewis2020retrieval}.  
To build a corpus of potentially relevant documents, we mined and crawled all \num{37982} clicked URLs in 
the web search engine's results
to the queries in $\mathcal{S}$. This includes but is not limited to the Azure subreddit URLs.
We discard the ones that require login, are behind a paywall, or are no longer available (broken links). To ensure that the URLs are of high quality, we also make sure they exist in Clueweb 2022 Set B \citep{clueweb22} top 200M most popular URLs. After filtering, we arrived at \num{23243} unique webpages. We used BeautifulSoup to convert each webpage's title and body into a plain text document, without any truncation.
The mean document length 
is \num{1246 \pm 1651} words (denoting mean $\pm$ standard deviation).

\subsection{Synthetic Dialogue Generation}
\label{sec:data:synthetic}
To generate synthetic dialogues in English of varying quality, we use 5 different LLM-based approaches (DS1--DS5), described in \cref{app:data_gen}.
These approaches have different levels of access to the document corpus.  Also, the true topic (which is always provided to the simulated user) is only revealed to the dialogue system in DS1--DS3.

We use \texttt{gpt-3.5-turbo-16k} with its default parameters \citep{openai2024} for all of our data generation (\cref{sec:data:synthetic}, \cref{sec:data:real}) and rubric-based evaluation (\cref{sec:exp_setup}).\looseness=-1

We randomly selected \num{50} topics, and used each of the systems DS1--DS5 to generate a synthetic conversation on that topic, resulting in \num{250} unique dialogues of varying quality. Each dialogue was evaluated by \num{3} judges (randomly assigned from a pool of \num{24} judges), resulting in \num{741}
personalized data points for dialogue evaluation after some guardrail quality checks (see \cref{app:data_gen_eval}). The average judge annotated \num[round-mode=uncertainty,round-precision=2]{30.95 \pm 13.02} dialogues. 

\subsection{Real Dialogue Collection and Evaluation}
\label{sec:data:real}
To obtain more realistic data for evaluation, we collect conversations with DS1--DS3 where the user turns are not generated by the LLM but by a real human.  The assistant in these three systems may be summarized as ``no RAG'' (DS1), ``oracle RAG based on the topic'' (DS2), and ``BM25 RAG based on the topic'' (DS3).\looseness=-1

The human who plays the user role in the dialogue then also serves as the judge for that dialogue, making them particularly well qualified to judge overall user satisfaction $Q_0$.
Details about the web interface and instructions to the humans can be found in \cref{app:interface}.

We collected a total of \num{223} evaluated human conversations by having 13 of the original 24 judges converse with systems DS1--DS3 (some judges were no longer available). Each judge engaged in and annotated \num[round-mode=uncertainty,round-precision=2]{17.15 \pm 3.41} dialogues on average.  The evaluations are summarized in \cref{app:real_data_eval}. 

\begin{table*}[t]
    \centering
    \resizebox{\textwidth}{!}{
    \begin{tabular}{llllll|llll}\hline
        & & \multicolumn{4}{c|}{\textbf{Synthetic Conversations}} & \multicolumn{4}{c}{\textbf{Real Human-Agent Conversations}} \\
          & \textbf{Model}  & \textbf{RMSE $\downarrow$} & \textbf{P's $\rho$ $\uparrow$} & \textbf{S's $\rho$ $\uparrow$} & \textbf{K's $\tau$ $\uparrow$} & \textbf{RMSE $\downarrow$} & \textbf{P's $\rho$ $\uparrow$} & \textbf{S's $\rho$ $\uparrow$} & \textbf{K's $\tau$ $\uparrow$} \\\hline
         1 & Random Eval  & 1.499 & 0.002 & -0.003 & -0.003 & 1.427 & 0.011 & 0.006 & 0.005 \\
         2 & Argmax LLM $Q_0$ & 0.984$^{1}$ & 0.153$^{1}$ & 0.161$^{1}$ & 0.147$^{1}$ & 1.186$^{1}$ & 0.106$^{1}$ & 0.123$^{1}$ & 0.120$^{1}$ \\
         3 & Expected LLM $Q_0$ & 0.856$^{12}$ & 0.182$^{1}$ & 0.217$^{1}$ & 0.168$^{1}$ & 0.901$^{12}$ & 0.143$^{1}$ & 0.141$^{1}$ & 0.138$^{1}$ \\
         4 & Calibrated LLM $Q_0$ & 0.801$^{123}$ & 0.198$^{12}$ & 0.196$^{1}$ & 0.193$^{12}$ & 0.784$^{123}$ & 0.211$^{123}$ & 0.218$^{123}$ & 0.192$^{123}$  \\  
         5 & FActScore \citep{min-etal-2023-factscore} & -- &  0.204$^{12}$ & 0.211$^{1}$ & 0.200$^{12}$ & -- & 0.216$^{123}$ & 0.218$^{123}$ & 0.207$^{123}$  \\
         6 & \llmeval  & 0.396\textsuperscript{1234e} & 0.401\textsuperscript{12345e} & 0.398\textsuperscript{12345e} & 0.393\textsuperscript{12345e} & 0.422\textsuperscript{1234} & 0.350\textsuperscript{12345} & 0.347\textsuperscript{12345} & 0.331\textsuperscript{12345}  \\\hdashline
        a & Oracle  & 0.237\textsuperscript{*bcdef} & 0.611\textsuperscript{*bcdef} & 0.626\textsuperscript{*bcdef} & 0.605\textsuperscript{*bcdef} & 0.289\textsuperscript{*bcd} & 0.717\textsuperscript{*bcd} & 0.711\textsuperscript{*bcd} & 0.675\textsuperscript{*bcd} \\
        b & \quad w/o LLM probs & 0.276\textsuperscript{*cef} & 0.551\textsuperscript{*cef} & 0.548\textsuperscript{*cef} & 0.533\textsuperscript{*cef} & 0.357\textsuperscript{*c} & 0.625\textsuperscript{*c} & 0.629\textsuperscript{*c} & 0.599\textsuperscript{*c}  \\
        c & \quad w/o Personalized Calibration & 0.401\textsuperscript{e} & 0.476\textsuperscript{*e} & 0.471\textsuperscript{*e} & 0.468\textsuperscript{*e} & 0.389\textsuperscript{*} & 0.582\textsuperscript{*} & 0.587\textsuperscript{*} & 0.565\textsuperscript{*}  \\
        d & \quad \quad $\drsh$ + Personalized isotonic regress & 0.273\textsuperscript{*cef} & 0.521\textsuperscript{*cef} & 0.526\textsuperscript{*cef} & 0.519\textsuperscript{*cef} & 0.302\textsuperscript{*bc} & 0.650\textsuperscript{*bc} & 0.653\textsuperscript{*bc} & 0.644\textsuperscript{*bc}  \\
        e & Depersonalized Oracle & 0.492 & 0.362 & 0.355 & 0.338 & -- & -- & -- & --  \\

         f & \quad \quad $\drsh$  + Personalized isotonic regress & 0.321\textsuperscript{*ce} & 0.482\textsuperscript{*e} & 0.485\textsuperscript{*e} & 0.477\textsuperscript{*e} & -- & -- & -- & --  \\
         \hline
    \end{tabular}
    }
    \caption{Performance on predicting human judges' $Q_0$ (overall quality). We report root mean squared error (RMSE) and, more important, correlations with human judges' responses (Pearson's $\rho$, Spearman's $\rho$, Kendall's $\tau$). 
    Results on the synthetic conversation dataset are based on 5-fold cross-evaluation; results on the real conversations are based on training on all synthetic conversations. 
    The superscripts denote statistically significant improvements according to a paired permutation significance test $(p < 0.05)$. The asterisk * means all methods in rows 1--6.}
    \label{tab:main_results}
    \vspace{-.5cm}
\end{table*}

\section{Experiments}
\label{sec:exp_setup}

We will evaluate how well \llmeval can predict individual judges' assessments $y^a_0$ of our $Q_0$ (overall user satisfaction).  We evaluate predictions $\yhat^a_0$ both in absolute terms (whether they achieve low root-mean-squared error, or RMSE) and in relative terms (how well $\yhat^a_0$ correlates with $y^a_0$, i.e., whether $\yhat^a_0$ can be used to rank $(T,a)$ pairs).

We train our calibration networks on synthetic dialogues.  We then evaluate them not only on held-out synthetic dialogues but also on real dialogues, to demonstrate that the LLM scoring and its calibration can generalize from synthetic to real data.

\paragraph{Hyperparameter Selection.}
To train a system on a given training set, we evaluate hyperparameter settings from a grid by 5-fold cross-validation on the training set, and then use the selected hyperparameters to train on the entire training set.  We select the hyperparameters that maximize the main task objective, namely the log-likelihood of (held-out) annotations $y_0^a$.  The hidden layer sizes $h_1, h_2$ each range over $\{10, 25, 50, 100\}$, the batch size ranges over $\{32, 64, 128, 256\}$, the learning rate of the Adam optimizer ranges over $\{0.00001, 0.00005, 0.0001, 0.0005, 0.001, 0.005,$ $0.01\}$, and the numbers of epochs for pre-training and fine-tuning each range over $\{5, 10, 20, 30, 40, 50\}$.\footnote{Instead of including the number of epochs in the hyperparameter grid search, an alternative would be to use a standard early stopping heuristic at each phase, by evaluating that phase's training objective periodically on held-out data.}   

\paragraph{Synthetic Data Evaluation.}
We test our calibration network on all 741 synthetic dialogues, using 5-fold cross-validation; the dataset is split at the dialogue level so that each dialogue appears in only one fold.  Different folds may select different evaluation hyperparameters, resulting in different architectures for the calibration network.\footnote{When training on 4 folds to evaluate the 5th, we select the hyperparameters by an inner 5-fold cross-validation on this training set of about 593 examples, as explained above.}

\paragraph{Real Data Evaluation.}
We test our calibration network on all 223 real dialogues, after training on all of the synthetic dialogues (again selecting hyperparameters by 5-fold cross-validation).

\paragraph{Baseline Methods.}
As \cref{tab:main_results} shows, we compare \llmeval to these \num{5} baselines:
\begin{enumerate}
    \itemsep0em 
    \item \textbf{Random.} For each dialogue independently, we produce 1, 2, 3, or 4 uniformly at random.

    \item \textbf{Argmax LLM $\boldsymbol{Q_0}$.} We use the top LLM prediction for $Q_0$: $\argmax_{y_0 \in \Y_0} \pllm(y_0 \mid T,  Q_0)$.  Note that this system always produces an integer.\footnote{In a pilot experiment, we found no significant improvement from few-shot prompting.}

    \item \textbf{Expected LLM $\boldsymbol{Q_0}$.} We use the expected value of the LLM's prediction for $Q_0$: $\sum_{y_0 \in \Y_0} y_0 \cdot \pllm(y_0 \mid T, Q_0) / Z_0$ (where $Z_0$ normalizes the probabilities over $\Y_0$---see \cref{fn:unnorm}).

    \item \textbf{Calibrated LLM $\boldsymbol{Q_0}$.} An ablated version of \llmeval that uses only $Q_0$, i.e., the feature vector $\vx = \left[\tilde{p}(y_0 \mid T, Q_0) \mid y_0 \in \Y_0 \right]$
    is restricted to the $Q_0$ answer probabilities. 
    We train and evaluate the calibration network just as for \llmeval, including cross-validation and hyperparameter selection.

    \item \textbf{FActScore} \citep{min-etal-2023-factscore}. This is a recent retrieval-based automatic evaluator\footnote{\url{https://github.com/shmsw25/FActScore}} that predicts the percentage of factually correct sentences as the overall evaluation score.  We use the Azure corpus described in \cref{sec:data:log_mining} as the retrieval corpus in FActScore, which performs better than the default Wikipedia corpus.

\end{enumerate}

\paragraph{Oracle Methods.}
    \Cref{tab:main_results} also shows upper bounds on performance.
    The \textbf{Oracle} system is the same as \llmeval, but the calibration network's input $\vx$---at both training and test time---includes the judge's \emph{actual} response to each question $Q_i$ (except for $Q_0$, which we aim to predict!\@) as a four-dimensional one-hot vector, in addition to the LLM response vector $\pllm(y_0 \mid T, Q_i)$.

    We ablate different components of the Oracle model by withholding the LLM response vector from the model input and by depersonalizing the  calibration network (\textbf{Oracle w/o Personalized Calibration}) by dropping $W_k^a$.  To restore a judge $a$'s idiosyncratic distribution of $Q_0$ scores (\cref{fig:judge_scores_dist}), 
  without restoring their idiosyncratic computation of $Q_0$ from other dimensions, we try correcting the output of the depersonalized calibration network using an $a$-specific isotonic regression model.  

 Our \textbf{Depersonalized Oracle} is similar to the Oracle, but instead of using the responses of the actual target judge $a$, it uses the distribution of responses of all other judges (averaging their one-hot vectors), holding out the target judge.\footnote{We cannot run this on the real conversation dataset, where each dialogue is annotated only by a single judge.}
 It also drops the personalized weights $W_k^a$.  

 Thus, the Oracle provides a rough upper bound on \llmeval.  The Depersonalized Oracle provides a rough upper bound on a version of \llmeval that produces $a$-independent results.

\section{Results}
\label{sec:results}

A trivial baseline of predicting a \emph{constant} $Q_0$ (the overall mean from training data) achieves an RMSE of 0.82 
on both synthetic and real conversations.  \llmeval roughly halves this (row 6 of \cref{tab:main_results}), so it explains $\approx \frac{3}{4}$ of the variance in human judgments of $Q_0$ across judges $a$ and texts $T$.  Its predictions of $Q_0$ have tolerably low error and correlate reasonably well with those of human judges.  

In sharp contrast, the LLM's direct response to $Q_0$ (row 2 or 3) does \emph{worse} than the constant baseline.  Even calibrating its response distribution for each judge (row 4) barely improves on the baseline, explaining only 5--10\% of the variance in human judgments and achieving only $\approx 0.2$ correlation with them.  This suggests that the LLM cannot help assess $Q_0$ (user satisfaction) until we ask it about the finer-grained dimensions $Q_1$--$Q_8$.

The results obtained by FActScore (row 5) do not correlate any better with overall satisfaction, so percentage of factually correct sentences is also not a good indicator of overall user satisfaction.  Moreover, \citet{liu-etal-2016-evaluate} showed that dialogue systems were poorly evaluated by simple metrics of lexical overlap with human responses.

\begin{table}[t]
    \centering
        \begin{tabular}{l@{ }ll}\hline
         \textbf{Model}  
            & \textbf{RMSE $\downarrow$} & \textbf{P's $\rho$ $\uparrow$}  \\\hline
        \llmeval & 0.422 & 0.350 \\ 
        \quad w/o fine-tuning & 0.493$^\triangledown$ & 0.249$^\triangledown$  \\ 
        \quad w/o pre-training & 0.525$^\triangledown$ & 0.226$^\triangledown$ \\
        \quad w/o personalization & 0.601$^\triangledown$ & 0.198$^\triangledown$ \\ 
        \hdashline
        \quad w/o $Q_0$ {\small(Satisfaction)} & 0.554$^\triangledown$	& 0.287$^\triangledown$ \\        
        \quad w/o $Q_1$ {\small(Naturalness)}  & 0.463$^\triangledown$	& 0.313$^\triangledown$ \\
        \quad w/o $Q_2$ {\small(Grounding Sources)} & 0.471$^\triangledown$	& 0.279$^\triangledown$ \\
        \quad w/o $Q_3$ {\small(Citation Presence)} & 0.573$^\triangledown$	& 0.075$^\triangledown$ \\
        \quad w/o $Q_4$ {\small(Citation Suitability)} & 0.497$^\triangledown$	& 0.311$^\triangledown$ \\
        \quad w/o $Q_5$ {\small(Citation Optimality)}  &	0.506$^\triangledown$	& 0.192$^\triangledown$ \\
        \quad w/o $Q_6$ {\small(Redundancy)} &	0.424	& 0.348 \\
        \quad w/o $Q_7$ {\small(Conciseness)} & 0.532$^\triangledown$ &	0.254$^\triangledown$	 \\
        \quad w/o $Q_8$ {\small(Efficiency)} &	0.510$^\triangledown$	& 0.161$^\triangledown$ \\
        \hline
    \end{tabular}
    \caption{Predicting $Q_0$: Ablation study on real conversation data for each design decision in our calibration network (top) and each rubric dimension (bottom).  $^\triangledown$~denotes a statistically significant performance drop from the full \llmeval ($p < 0.05$).}
    \label{tab:main_ablation}
    \vspace{-.5cm}
\end{table}

\begin{table}[t]
    \centering
        \begin{tabular}{lll|ll}\hline

           & \multicolumn{2}{c|}{\textbf{Expected LLM $Q_i$}} & \multicolumn{2}{c}{\textbf{\llmeval}} \\
        &  \textbf{RMSE $\downarrow$} & \textbf{P's $\rho$ $\uparrow$} &  \textbf{RMSE $\downarrow$} & \textbf{P's $\rho$ $\uparrow$} \\\hline
          $Q_0$ & 0.901 & 0.143 & 0.422$^*$ & 0.350$^*$ \\  
           $Q_1$ &1.033 & 0.177 & 0.637$^*$ & 0.318$^*$ \\
           $Q_2$ &  0.799 & 0.140 & 0.543$^*$ & 0.265$^*$  \\ 
            $Q_3$ &  0.796 & 0.347 & 0.532$^*$ & 0.511$^*$ \\  
           $Q_4$ &  0.919 & 0.166 & 0.706$^*$ & 0.494$^*$ \\ 
           $Q_5$ &  1.104 & 0.191 & 0.786$^*$ & 0.387$^*$    \\ 
            $Q_6$ & 1.726 & 0.030 & 0.430$^*$ & 0.279$^*$ \\ 
            $Q_7$ & 1.240 & 0.057 & 0.693$^*$ & 0.318$^*$ \\ 
            $Q_8$ & 0.981 & 0.059 & 0.232$^*$ & 0.249$^*$ \\ 
        \hline
    \end{tabular}
    \caption{How well can \llmeval predict the response $y_i$ to question $Q_i$? For each row, we fine-tune \llmeval \emph{on the target rubric dimension} and compare to  Expected LLM $Q_i$ on the real conversation data.  Superscript $^*$ indicates statistically significant improvement with 95\% confidence ($p < 0.05$).}
    \label{tab:per_dim_results}
    \vspace{-.3cm}
\end{table}

\section{Analysis}
\label{sec:analysis}

\paragraph{Calibration.} Does our trained \llmeval produce well-calibrated probability distributions for $Q_0$ (as one would expect from maximum-likelihood training)?  We checked on synthetic data. It obtained excellent smECE values of $< 0.05$ for each $y_0 \in \Y_0 = \{1,2,3,4\}$, where smECE is the smoothed expected calibration error \cite{blasiok2023smooth}.
Informally, this means that for each $y_0 \in \Y_0$, when we examine the held-out examples $(T,0,a,y_0^a)$ with $\phat_a(y_0 \mid T, Q_0)\approx p$, the fraction where $y_0^a=y_0$ was in fact $\approx p$.  \Cref{app:calibration} shows calibration plots and discusses how to use calibrated probabilities for downstream decisions.

\paragraph{Ablation Studies.}

\Cref{sec:results} showed that LLM responses on \num{8} additional questions were useful, but was our calibration network the best way to incorporate them into our prediction of $Q_0$?  To justify each design decision, we try omitting pre-training, fine-tuning, and personalized weighting from our calibration network. The results on the real conversation data in \cref{tab:main_ablation} show that predictions were improved by each step.  In particular, it was indeed useful to do multi-task pre-training of the calibration network (which required human judgments on all questions) and to then fine-tune on the main task.  Personalized weighting had the greatest impact.

Also, were all \num{8} questions useful?  We measured the impact of each question by omitting it from the evaluation rubric for the \llmeval model (bottom half of \cref{tab:main_ablation}). All rubric dimensions contributed significantly to the $Q_0$ prediction, except for $Q_6$, which focuses on redundancy in the dialogue.  
Using even more rubric dimensions might improve performance further (\cref{fn:differentq,app:extensions}). That said, considering more rubric dimensions would mean more human annotations at pre-training time and/or more LLM computation.

\paragraph{Oracle study.}
Giving \llmeval access to 
a judge's true responses to $Q_1$--$Q_8$ lets us see how well the judge's overall quality score $Q_0$ 
is predictable from our particular rubric dimensions.  This gets rather better results, including an excellent 0.72 Pearson's $\rho$ correlation between predicted and actual satisfaction scores on real dialogues (row `a' in \cref{tab:main_results}).  Almost all of this performance can be obtained from \emph{only} the judge's responses, without access to the $\pllm$ score distributions (row `b').\looseness=-1

This suggests a future strategy (discussed below) of improving the \emph{input} to \llmeval by getting the LLM to better predict the judge-specific human rubric responses that were available to the Oracle (row `a'), or at least judge-independent versions (rows `e'--`f').
Once such responses are available, the ensembling is still best done by a calibration network that understands an individual judge's preferences---though under oracle conditions and with our population of judges, dropping that personalization would not be dramatically worse (row `c'), and a fraction of the difference can be made up simply by adjusting the predicted scores $\yhat_0$ with personalized isotonic regression (row `d').
\looseness=-1

\paragraph{On which dimensions do zero-shot LLMs need improvement?}  \Cref{tab:per_dim_results} shows these results. 
  Redundancy ($Q_6$), Conciseness ($Q_7$), and Efficiency ($Q_8$) were especially difficult for the LLM to predict---it showed close to zero correlation with human annotators.  \llmeval much better predicted these scores, as well as overall Satisfaction $Q_0$, by exploiting the full response vector $\vx$: e.g., it improved RMSE by $> 0.5$ in all of these cases.

  The LLM's errors on a difficult question $Q_i$ could potentially be reduced through prompt engineering, few-shot prompting, fine-tuning the LLM, or calling a larger LLM.  
  Is that worth it? 
  To assess the potential improvement to $Q_0$ prediction from better answering $Q_i$, one could use cross-validation to evaluate the benefit to $Q_0$ from replacing just $Q_i$ with oracle scores before training \llmeval.

\paragraph{How much human judge data is needed to train calibration?}  See \cref{app:learning_curves} for learning curves.

\section{Related Work}\label{sec:related}

\paragraph{LLM Evaluation} Zero-shot or few-shot LLM evaluators
have been shown to have higher agreement with human annotators than traditional lexical overlap or even earlier transformer embedding models, across a variety of natural language generation (NLG) tasks \citep{fu2023gptscore,lin2024interpretable}. Furthermore, when compared to crowdworkers, LLMs can have higher agreement with expert annotators \cite{Gilardi2023ChatGPTOC,chiang-lee-2023-large}. Additional techniques like chain-of-thought prompting and auto-prompt engineering can also further improve alignment with human ground truth \citep{liu-etal-2023-g,liu2023CalibratingLE,lin2024interpretable}. It seems that LLMs are capable of measuring an increasing range of evaluation dimensions including factuality \citep{min-etal-2023-factscore, gekhman-etal-2023-trueteacher,yue-etal-2023-automatic}, interpretability \citep{Lu2023ErrorAP}, and relevance  \citep{saadfalcon2023ares}. These works generally focus on average judge preferences on individual evaluation attributes, while we focus on using LLMs to capture the interplay of individual attributes to better predict all judgments (particularly of overall text quality) for a given judge.\looseness=-1

\paragraph{Calibration of LLM evaluators.}

\citet{zhao2023pareto} develop a Pareto-optimal method for estimating the error rate of an LLM-based predictor by combining both LLM and heuristic predictions, which can in turn be used to correct the initial
LLM prediction. While they similarly take advantage of an ensemble of predictors, they assume specific ground-truth answers, whereas \llmeval produces distributions over reasonable answers.

\paragraph{Subjectivity in Evaluation.} While LLMs can agree with expert judges, in cases where experts have low agreement, LLMs tend to have low agreement with the judges as well \cite{chiang-lee-2023-large}.  It is increasingly acknowledged that accounting for subjectivity (as opposed to collapsing or removing disagreements) in NLP evaluation is a key part of evaluation design \cite{pavlick-kwiatkowski-2019-inherent,basile-etal-2021-need,uma-etal-2021-semeval,uma2021disagreement,plank-2022-problem,plepi-etal-2022-unifying,sandri-etal-2023-dont}.
By training a single network to model all judges, we take the view that ``disagreement is not noise but signal'' \cite{aroyo-welty-2015}.  \Citet{baan-etal-2022-stop} put it more starkly: without modeling the judge distribution, metric calibration is itself nonsensical on subjective tasks.  Making downstream use of these disagreeing judges---or rather \llmeval's simulation of them on new texts---is discussed by \cref{app:dashboard}, \citet{gantt-etal-2020-natural}, and \citet{uma2021disagreement}.

While our work is similar conceptually to \citet{gantt-etal-2020-natural} in that we include judge-specific parameters to predict each human judge's responses, we show that this can be further improved by predicting responses to multiple questions (our auxiliary tasks $Q_1$--$Q_8$ along with our main task $Q_0$).

\citet{xiao-etal-2023-evaluating-evaluation} analyze common NLG evaluation dimensions and metrics using the concepts of reliability and validity from measurement theory. They find that while manual judges may rate generated texts by different dimensions like `coherence' or `relevance,' these dimensions can exhibit poor validity structure. In their case, this means that they find that an individual judge's correlation with their own ratings across coherence and relevance can be as high or higher than correlation between other judges within each dimension, supporting the idea individual judges may have idiosyncratic or conflated mappings of different evaluation criteria. \citet{xiao-etal-2023-evaluating-evaluation} suggest several ways to improve the dimensions to account for this. We did not preform a similar analysis on our judges and rubric dimensions, although improvements here would be orthogonal to the benefits of \llmeval, since judges 
may reasonably disagree even in the absence of validity structure issues.  

\section{Conclusions}\label{sec:future}

This work proposed \llmeval---a rubric-based framework for automatic evaluation of text.  We trained and tested it on novel datasets of information-seeking dialogues.
\llmeval performs multidimensional evaluation using a black-box LLM, then aggregates and calibrates these multidimensional responses for each human judge. 

Although the LLM's raw responses do not highly correlate with human judgments in such a complex task, we found that combining its response distributions on all questions can predict each human judge's responses, including overall satisfaction.  We obtained substantial improvements on RMSE and on both linear and rank-based correlation metrics, on held-out synthetic conversations (development data) and real ones (test data).  Below, we discuss limitations, ethics, uses, and extensions.

\section*{Acknowledgments}
We thank Val Ramirez and the 
data specialists who contributed to the
creation of this work.

\section*{Limitations}

\paragraph{Robustness.} In general, one might hope that the trained \llmeval can successfully predict human scores even in \emph{new} test domains---at least when it is given a broadly competent LLM, a broadly worded rubric, and training examples that exhibit a variety of score profiles on that rubric.  However, we did not evaluate this, other than showing that our trained \llmeval worked well when applied to a \emph{slightly} different test distribution (real rather than synthetic dialogues) on the same topics (information-seeking Azure queries). 

Robustness is particularly important when privacy rules prevent having human judges evaluate real examples from the test distribution, as in some deployed dialogue systems or when coding medically or legally sensitive data.  Even when training examples can be drawn from the true test distribution, it may be hard to find judges who are competent to annotate the full range of topics and styles in such examples.  For example, judges may be unavailable for low-resource languages---and it is not necessarily true that LLM scores bear the same relation to human scores for texts in those languages, since the LLM may be less competent to judge such texts \citep{ahuja2024benchmarking}, or the texts themselves may have different quality issues.\footnote{For example, when a multilingual dialogue system is used in a low-resource language, user satisfaction $Q_0$ may be lower because of language-specific problems such as formality that did not arise in \llmeval's training, or were not as highly weighted, or were not directly assessed by the rubric at all.}

Robustness is also needed when the test distribution shifts over time---either for exogenous reasons such as new topics or user populations, or because the metric has become a target (Goodhart's Law) so that the texts are increasingly designed to score well on predicted $Q_0$.  The latter case includes NLG engineering, as well as adversarial settings like essay grading or legal discovery, where test-takers or email conspirators have an incentive to write their texts so as to fool the evaluation system.

\paragraph{Efficiency.} We used a large pretrained LLM to answer each rubric question.  
It would be cheaper to use smaller models where possible, perhaps fine-tuned on specific questions.
One could also decide which questions are worth asking (and which models to ask) by using an \emph{adaptive} rubric: e.g., choose the next evaluation question to maximize the expected information gain, and stop at the point of diminishing returns, so that it is not necessary to ask all questions.  An adaptive rubric could in principle be quite large, with only a small portion of it used on any particular text $T$.  This direction and other possible extensions are discussed in \cref{app:extensions}, but we did not try them.

\paragraph{Downstream Evaluation.} Although we measured overall correlation between predicted and human scores on each rubric question, we did not evaluate the usefulness of our predicted scores for difficult downstream tasks such as choosing among similar candidate answers or dialogue systems.  More rubric questions might be needed for sufficiently accurate evaluation (see \cref{fn:differentq,app:extensions}). 

A particularly challenging but important downstream use is to \emph{improve} natural language generation.  We have not addressed this.  However, a metric such as our predicted overall quality $\yhat_0$ (averaged over a set of judges as in \cref{app:agg}) could be used as a reward signal, for example to improve an LLM by proximal policy optimization \citep{Schulman2017ProximalPO}.  More ambitiously, one could train the LLM using multi-objective reinforcement learning 
\citep[e.g.,][]{yang2019generalized,abels2019dynamic,rame2023rewarded,wu2023fine}
to consider idiosyncratic preferences \emph{at runtime} and generate text that achieves a high predicted \emph{user-specific} reward.  For example, one could use $\yhat_0^a$ as the runtime reward function if one modified our calibration network to do regression (\cref{sec:method}) via $\yhat_i^a = (\mathbf{v}_i+\mathbf{v}_i^a) \cdot [1;\vz_2]$ where $\vz_2$ is judge-independent (compare \cref{eq:softmax}).  Then 
$\vz_2$ serves as a multi-objective reward vector, and $\mathbf{v}_0+\mathbf{v}_0^a$ is the preference weighting that linearly scalarizes this reward at runtime, where $\mathbf{v}_0^a$ may be regarded as a preference embedding of the user $a$ (possibly computed from features of $a$).

\paragraph{Fine-Grained Evaluation.} We only considered evaluating entire texts.  However, humans often perform finer-grained evaluation tasks---such as highlighting \emph{problematic} spans in human- or machine-written text (e.g., to provide feedback and opportunities for revision), or highlighting \emph{relevant} spans (e.g., to call a human or machine's attention to them).  We have not presented methods for automating or calibrating fine-grained evaluation.

\section*{Ethics Statement}

\paragraph{Beyond User Satisfaction.} Evaluation metrics drive engineering and so have real-world consequences.  Our experiments focused on predicting overall user satisfaction (our choice of $Q_0$), but we do not in fact recommend this as the actual goal of dialogue system development.  In practice, quality evaluation of a dialogue agent should also assess potential harms to the user (e.g., false information), to the dialogue system owner (e.g., reputational harm through violating policies on content or style), and to third parties (e.g., encouraging violence or outputting private or copyrighted information).

\paragraph{Fairness Auditing.} 
Our aligned LLM’s ability to approximately match human judges does not answer the question of whether the unaligned LLM, the aligned LLM, the human judges, or the manually constructed rubrics are fair or unbiased. 
Even when our system does achieve low total error at matching fair judgments, it is not guaranteed that its errors or their downstream harms are evenly distributed.  Thus, accuracy (\cref{tab:main_results}), calibration (\cref{app:calibration}), and rubric validity should be checked for various important subsets of the data.  For example, in essay grading, does the calibrated LLM systematically underestimate the quality of the ideas of speakers of a particular dialect?  In dialogue system evaluation, is a particular user population frustrated with a certain kind of error that they experience heavily, yet this type of error is underdiagnosed?\footnote{Or, going beyond auditing, one could try to learn a \emph{multicalibrated} model in the first place \cite{hebertjohnson2018multicalibration}.  Such a model's \emph{average} rating over a subset of texts $S$ will be approximately correct, for \emph{every $S$ in a given large family} of subsets that are computationally identifiable and not too small.  This ensures that the errors are in a sense fairly distributed: the model cannot systematically underestimate or overestimate texts written by any particular subpopulation of authors, preferred by particular judges, having particular linguistic features, etc.  Typically, a multicalibration algorithm builds up a complex model (without sacrificing accuracy): each step augments the current model with a learned post-correction step that adjusts the outputs on some subset of inputs.  Such algorithms exist for regression \cite[e.g.,][]{globusharris2023multicalibration} as well as classification, and have recently been applied to LLM evaluation \cite{detomasso2024multicalibration}.}

\paragraph{Human Data.} \llmeval requires collecting data from human judges that reveal their personal preferences, such as their affinity for specific textual passages.  Such data should always be carefully safeguarded.  In certain cases it may even be appropriate to train the calibration network using differential privacy, to make it impossible to guess information about particular judges from the network weights.

\paragraph{Harmful Uses.} \llmeval may enable generating or choosing content that appeals to a specific human's preferences. This could improve their satisfaction with the NLG output, but it could also be used to optimize for their engagement---even when this is harmful (for example, confirming biases, spreading misinformation, provoking outrage, swindling, or recommending antisocial actions or self-harm).

\paragraph{Environmental Costs.} \llmeval is compute-intensive, as it involves calling an LLM several times for each NLG output. On a small evaluation dataset, the compute cost may be modest, but \llmeval will add to the environmental footprint of a system if it is applied to a substantial fraction of user traffic, or is called many times during a hyperparameter tuning loop or to compute the reward signal for reinforcement learning.  Costs might be reduced through distillation or an adaptive rubric, as discussed in the Limitations section.%

\bibliography{acl2024}

\clearpage\appendix\onecolumn

\section{Aggregating Predicted Scores}\label{app:agg}\label{app:dashboard}

Our use of judge-specific distributions $\phat_a$ can be regarded principally as a technique to improve training on human annotations.  Judges are heterogeneous (\cref{fig:judge_scores_dist}) and a training document will only be judged by some of them (\cref{sec:data}), as discussed in \cref{app:extensions} below.  Knowing who the judges were can help us model the training data.  For example, suppose $T$ got mostly low LLM scores, yet all judges randomly assigned to $T$ in training data gave it a high overall score.  The model might ``explain away'' the high scores if it knows that those particular judges are generous or are focused on dimensions where $T$ did well---and thus could still predict low overall scores from the remaining judges.

However, this means that our trained calibration network does not produce ground truth.
It only models the idiosyncrasies of individual judges $a$ (successfully, as shown in \cref{fig:judge_scores_dist,tab:main_results}). We do not even suggest that purely objective scores exist (see \cref{sec:related}), except on extremely precise rubric questions.  So which judge should we use in the end?  That is, after training \llmeval, how should we practically obtain a final assessment of a new text $T$?

We might use the mean predicted overall quality, $\yhat_0$, where $\yhat_i=\mean_{a \in \A}\yhat^a_i$ for a fixed set $\A$ of \emph{trusted judges}.\footnote{Any judges not in $\A$ still help regularize the training.  They might be omitted during fine-tuning (just as $Q_i$ was for $i\neq 0$).}  This assumes that $Q_0$ calls for numerical responses on an interval scale (see \cref{fn:interval}), so that the mean is defined and meaningful.  An unweighted mean also assumes that we equally want to please all judges in $\A$ (see the start of \cref{sec:method}).  The benefit of \llmeval is that we do not actually query these judges---we predict how each of them \emph{would} respond by querying an LLM and calibrating \emph{its} response distributions.

What makes a judge ``trusted''?  The judges in $\A$ might have had additional training, insight, information, or time.  For example, \citet{thomas2024large} distinguish between trained assessors and third-party crowdworkers.
If \llmeval scores are used to nominate interesting documents for more careful manual review, for example in a legal document review workflow, then $\A$ might consist of the experienced lawyers or paralegals who perform the manual review (and who will continue to add to the training set by answering at least $Q_0$ on newly nominated documents).  Alternatively, a trusted judge, rather than being a single human, might correspond to the result of a discussion and reconciliation process among multiple untrusted human judges.

The various applications in \cref{sec:intro} might call for other ways to aggregate the predicted judgments (or the resulting document rankings).  E.g., to be safe, lawyers may want to replace $\mean$ with $\max$ in the definition of $\yhat_0$ to review any document that at least one judge in $\A$ would have deemed relevant.  The predicted judgments can also be used \emph{without} aggregation \cite{uma2021disagreement,plank-2022-problem,gantt-etal-2022-decomposing} to train or evaluate other systems for generating or scoring text.

\paragraph{Dashboards.} 
In our setting of dialogue evaluation (or NLG evaluation), the mean predicted score $\yhat_0$ for a given text $T$ can be used as a target metric for system development and monitoring.  

To aid system developers, we can go beyond $\yhat_0$ and compute $\yhat_i$ on $T$ for \emph{each} $Q_i$ (using a version of the network that has been re-fine-tuned to predict $\yhat_i^a$ as its main task).  We can also quantify the importance of improving $T$ to raise its mean human $Q_i$ rating: $\frac{\nabla_{\vx} \yhat_0 \cdot \nabla_{\vx} \yhat_i}{\nabla_{\vx} \yhat_i \cdot \nabla_{\vx} \yhat_i}$ estimates the improvement in the prediction $\yhat_0$ per unit of improvement in the prediction $\yhat_i$, if one could change $T$ so as to change $\vx$ in the direction of steepest ascent of $\yhat_i$.\footnote{Of course, it will not usually be possible to change $T$ in quite this way: the desired direction $\nabla_{\vx} \yhat_i$ may send $\vx$ out of the feasible space of texts.  Thus, a more sophisticated approach is to estimate the manifold of
plausible $\vx$ vectors from known training texts (including desirable texts), so that each $\vx$ can be represented in terms of underlying manifold coordinates $\vw$ and residuals.  Now $\nabla_\vx$ may be replaced with $\nabla_\vw$ throughout. 
 This constrains the steepest-ascent direction to point along the manifold.  The manifold may be estimated with methods such as Isomap \cite{tenenbaum2000global}, LLE \cite{roweis2000nonlinear}, or VAE \cite{kingma-welling-2019}.  Less ambitiously, one could merely represent the $\pllm$ distributions within $\vx$ using softmax parameters $\vw$, so that steepest-ascent using $\nabla_\vw$ will at least constrain these distributions to the probability simplex.}

A dashboard for the system developers could show how all of the above quantities are distributed over a representative set of texts $\T$, using kernel density estimation (or a histogram).
The dashboard could also \emph{display} these distributions for different subsets of $\T$ representing specific topics or groups of users, could \emph{compare} them across different versions of the system, and could \emph{track their means or quantiles} over time.  Uncertainty bands around each density curve can be found by computing it many times, each time substituting bootstrap replicates of $\A$ and $\T$ 
and---in the case of the density of $\yhat_i$---replacing each $\yhat_i^a$ for each text $T$ with a sample from $\phat_a(y_i \mid T,Q_i)$.\footnote{As a caveat, this procedure assumes that the true $y_i^a$ values actually follow this joint distribution, i.e., that the calibration network is correct.  To incorporate uncertainty about the network's parameters as well, we would also have to retrain them each time on a bootstrap replicate of the training set.  Then a small training set would also lead to wider uncertainty bands.  We would also likely get wider uncertainty bands by modeling and sampling the judgments $y_i^a$ jointly (for each $i$). We currently model them as independent (see \cref{fn:nonindep}), 
but this assumes that the errors $\yhat_i^a - y_i^a$ are uncorrelated.  In fact they are likely to be positively correlated across judges $a$ on the same text $T$ and also across similar texts, since they are derived from the same or similar LLM response vectors $\vx$.}
Thus, small $\A$, small $\T$, and high-variance distributions $\phat_a$ for $a\in\A$ will all lead to wider uncertainty bands.  This procedure also yields confidence intervals on the statistics (means, differences of means, etc.).

Each of the above distributions over $\T$ could optionally be disaggregated into a distribution over $\T \times \A$.  
Suppose $\Y_i$ is a 1--4 Likert scale of ``strongly disagree, disagree, agree, strongly agree'' and $|\A|=2$.  If one judge probably disagrees and the other probably strongly agrees with $Q_i$ for a given text ($\yhat_i^a \approx 2.0$, $\yhat_i^{a'} \approx 4.0$), then these two opinions would be recorded separately in the disaggregated view, rather than being averaged into ``agree'' ($\yhat_i \approx 3.0$).  Averaging Likert responses is often discouraged because it homogenizes diverse opinions and because it treats the Likert scale as an interval scale rather than an ordinal scale \cite{barry2017averages}.\footnote{Disaggregation therefore avoids averaging over judges.  Even then, however, each $\yhat_i^a$ is still itself a weighted average over possible responses by $a$.  This inner average may be problematic as well (\cref{fn:interval}).  Still, it elides only uncertainty, not disagreement, so disaggregating it seems less useful.}
We suspect, however, that both aggregated and disaggregated views are useful in practice.  Clicking on the lower tail of an \emph{aggregated} distribution will display problematic dialogues that are predicted to have a low \emph{average} score on $Q_i$.  For a \emph{disaggregated} distribution, the same click displays dialogues that are predicted to be problematic for \emph{specific} judges, according to their idiosyncratic interpretations of $Q_i$.

\section{Handling Other Types of Datasets}\label{app:extensions}

Our experimental datasets used a certain kind of rubric and a simple data collection mechanism.  However, the idea of predicting human judgments with a calibration network is quite general and can be extended to a variety of practical settings.  We discuss some useful examples in this appendix.

\paragraph{Additional Features.}  Our calibration network's input is only $\vx$, the vector of LLM responses on $T$.  To give it access to other predictive features, $\vx$ or $\vz_2$ could also be augmented with a fixed-dimensional embedding of $T$ (as already noted in \cref{fn:embedding}).  The embedding function could be pretrained, or it could be fine-tuned jointly with the calibration network.

To avoid overfitting if the embeddings are high-dimensional, the embeddings can be replaced with $\mathbf{0}$ during initial training.  When the embeddings are revealed in the next phase of training, they may reveal properties of the text that systematically cause the calibrated LLM to overestimate or underestimate the human judges' answers to certain questions.  The calibration network can then learn to use them to further correct its estimates. 

This is an example of the general principle that regression can be more accurate with more regressors.  For the same reason, it may be useful for $\vx$ to include additional LLM questions (see \cref{fn:differentq}), which might cover additional criteria or use variant prompts.  Ambitious questions might potentially ask the LLM to think step-by-step about the user's goals and whether they are achieved (chain of thought), or to take on specific personas \cite{wang2024unleashing} that might reflect the values and needs of some human judges.
If internal states of the (Transformer) LLM are available, $\vx$ can be further enriched with information about how it computed each distribution $\pllm(y_i \mid T,Q_i)$, such as a high-layer encoding of the final token of the $T, Q_i$ prompt, which strongly influences this distribution.\footnote{Thanks to Zhichu (Brian) Lu for this observation.} Similarly, $\vx$ could include other features of $T$ that are extracted by manual code or trained classifiers rather than by prompted LLMs.  It could even include features of the judge $a$, which allows sharing parameters across similar judges---especially useful when end users are enlisted as judges (discussed later in \cref{app:extensions}).
Finally, it may improve accuracy to include available \emph{metadata} about $T$, such as its domain, date, and author---but such metadata should be masked for predictions that will be used to compare performance on different domains, dates, or authors, so that the predicted scores are fair in the sense that they depend only on the text $T$.

\paragraph{Missing Features.} The Limitations section suggested using an ``adaptive rubric'' to reduce the number of queries to the LLM at test time.  An adaptive rubric would query the LLM dynamically to ask the most useful questions first and to ask only as many questions as are needed to predict target quantities such as $\yhat_0$.\looseness=-1

However, this requires being able to predict $y_i^a$ values even when some of $\vx$ is missing.\footnote{We can represent a missing LLM response in $\vx$ by having  $\pllm(y_i \mid T, Q_i)$ put all of its probability on a special value $y_i=\textsc{mask}\notin\Y_i$.}  If we train \llmeval with dropout, then it will be able to handle this case.  

Furthermore, we can extend the \llmeval output so that it predicts not only distributions over the human responses $y_i^a$, but also distributions over the missing parts of $\vx$ \cite{uria2016neural,devlin-et-al-2018,kachuee2019dynamic,covert2023learning}---that is, over what the LLM might say if asked.  This can be used for dynamically choosing the next LLM question.  Dynamic feature selection dates back at least to \citet{he-daume-eisner-2012-icmlw}.  We envision an approach similar to that of \citet{covert2023learning} and \citet{zhong-et-al-2023}, which greedily selects the next LLM question $Q_i$ based on information gain---essentially, based on how much the variance of the predicted $\yhat_0$, for example, is expected to decrease after observing the LLM's distributional answer to $Q_i$, $\pllm(y_i \mid T, Q_i)$. Computing this requires us to guess how the LLM is likely to respond to $Q_i$, 
given its responses to previous questions (i.e., we consider how it might fill in the missing part of $\vx$ given the part of $\vx$ that has been observed so far, and average over these possibilities).

Dealing with missing features is also necessary if the input feature set evolves over time.  We may not wish to compute old features on new data, or new features on old data.  Indeed, we may not be able to do so, if the feature has changed because the underlying LLM has been replaced with a new version.

\paragraph{Irregular Datasets.}
Our training objective \labelcref{eq:loglik} tries to predict $y_i^a$ for each $(T,i,a,y_i^a)$ tuple in the training dataset $\dataset$.  Any collection of tuples can be used.
That is, it is not necessary to obtain answers to all human questions for every text, or to use the same judge for all questions on a text.  This is often useful.

First, in practice, new texts $T$ or questions $Q_i$ may periodically be added to the training dataset $\dataset$ to better cover the observed distribution of test data and to track newly identified issues.  The set of judges may also change over time due to staff turnover.  As new tuples are collected, \llmeval can simply be retrained on the growing, heterogeneous dataset.

Second, perhaps not every question is applicable to every text, and not every human judge has the same expertise.  Thus, each text $T$ might select a different set of questions $Q_i$, and might route different questions to different judges.  A manually written policy can rule out inapplicable questions (for both human judges and LLM evaluators) by consulting text classifiers or the results of earlier questions.  The applicable questions should be routed to judges with appropriate expertise,\footnote{We remark that to complement the judges' own expertise, one might equip them with information beyond $T$.  That is, for some $(a,i)$ pairs, the judge $a$ could consistently be shown additional information, such as the output of a fact-checker or machine translation system, or the user's reaction to the system response.  The trusted judges $\A$ of overall quality ($Q_0$) could be shown expert judges' responses to other rubric questions, or their response distributions as predicted by \llmeval.}  which---depending on the question---may include familiarity with $T$'s topic, dialect, or type of user.  \Citet{yoshikawa2021tell} review and propose methods for routing texts to judges---a problem that is closely related to dynamic feature selection above.  Some questions may require special expertise independent of $T$, e.g., questions that assess the harmfulness or inappropriateness of a dialogue system's response according to the policies of the system's owner.

Third, even when it is reasonable to ask a particular question of a particular judge, doing so may not be the best use of a limited annotation budget.  One may use an active learning workflow \cite{settles-2012} that prioritizes annotations that are not already predictable by \llmeval---that is, where $\phat_a(y_i \mid T,Q_i)$ still has high variance after $\phat_a$ has been trained on previously collected data.

Fourth, in a dialogue system, we may be able to enlist our end users as additional judges, which is especially useful on private dialogues that only the users can see. For example, it is currently common to give users the opportunity to click {\faThumbsOUp} or {\faThumbsODown} (which may in turn trigger followup questions).  We regard this click or non-click as just another human judgment $y_i$ that we wish to predict.\footnote{Similarly, we may treat ``Did the user choose to visit the system again the next day?'' or ``How long before the user's next visit?'' as a more implicit human judgment $y_i$ that we wish to predict.}  Note that this question is distinct from the question $Q_0$ that asks for the overall quality of the text (which usually uses a Likert scale and which may ask about aspects of the dialogue beyond user satisfaction).  The calibration network can be used to predict the user's response---that is, a click or non-click\footnote{``No click'' will usually have probability close to 1.  To avoid a large number of low-information training examples, one can downsample the ``no click'' examples in training data from this domain, provided that $\vx$ indicates whether the example comes from a downsampled domain (since this kind of downsampling will shift the priors on many questions $y_i$ toward more extreme responses, and thus should shift the hidden features $\vz_1,\vz_2$ guessed from an ambiguous example $\vx$).  Also, to control the number of parameters in a system with many users, a reasonable simplification is to fix $W_k^a = 0$ when $a$ is a user. Then for each user $a$, we only have to learn a matrix $V_i^a$ with two non-zero rows (for {\faThumbsOUp} and {\faThumbsODown}; the row for no response can be fixed at $\mathbf{0}$, WLOG).   Note that in the common case where user $a$ has never provided any explicit feedback, so that $V_i^a=0$, the backoff matrix $V_i$ still ensures a reasonable prediction---particularly if $a$'s demographics and/or user behavior are represented in $\vx$ when predicting the answer to this question, allowing the network to share statistical strength with similar users.
}%
---from various LLM questions.
Some of these LLM questions may be designed to detect various kinds of verbal feedback from the user, such as praise or specific complaints, rather than assessing the system's responses directly.  In fact, \citet{lin2024interpretable} present a method for \emph{automatically} creating questions of this sort from private dialogues.  Questions about verbal feedback may also be presented to additional human judges---though only on synthetic or other non-private dialogues---so that they contribute to multi-task regularization of \llmeval and so that calibrated versions can be shown on a dashboard (\cref{app:dashboard}).

\paragraph{Heterogeneous Response Types.}
\Cref{eq:softmax} constructed a softmax distribution $\phat_a$ over a small finite response set $\Y_i$.  But if some $Q_i$ demands real-valued responses (e.g., $\Y_i=\mathbb{R}$), then $\phat_a(y_i \mid T, Q_i)$ for that $i$ can simply be changed to a density model, where the calibration network predicts the parameters of some parametric density function from $\vz_2$.  Similarly, if some $Q_i$ demands textual responses (e.g., $\Y_i=\Sigma^*$), then $\phat_a(y_i \mid T, Q_i)$ can be changed to an autoregressive language model conditioned on $\vz_2$.

Next, consider the case where $\Y_i$ is finite but large.  Here the matrices in \cref{eq:softmax} are large, so generalization might be improved by smoothing them.  This can be done by parameterizing $V_i = Y_i U_i$ and $V_i = Y_i U_i^a$, where the rows of matrix $Y_i$ serve as embeddings of the various responses $y_i \in \Y_i$.  Similar responses should have similar embeddings.  The unsmoothed case takes $Y_i$ to be the identity matrix (yielding one-hot embeddings), but using a learned matrix $Y_i$ with fewer columns can reduce the number of parameters.  In some cases, $Y_i$ does not even need to be learned: pre-trained word embeddings can be used if $\Y_i$ is a natural-language vocabulary, and systematic number embeddings \cite{gorishniy2022numerical} can be used if (e.g.) $\Y_i=\{1,\ldots,100\}$.  

The preceding paragraph treats large response sets for human judges, which are predicted by the \emph{output} of the calibration network.  If the LLMs are also permitted to use large response sets, which appear in the \emph{input} of the calibration network, a similar solution applies: premultiply the vector $\pllm(y_i \mid T, Q_i)$ by $Y_i^\top$ to reduce its dimensionality before including it in $\vx$.  For infinite response sets as in the first paragraph, standard schemes can be used to embed numbers \cite{gorishniy2022numerical} or text \cite{devlin-et-al-2018}.

Finally, returning to the setting of our own experiments, we observe that when $\Y_i$ is an ordinal scale with $n$ possible responses, 
such as a Likert scale, it is not strictly necessary to use a flexible softmax distribution as we did in \cref{eq:softmax}.  Instead, 
$y_i^a$ could be modeled with fewer parameters as a quantized version of an \emph{underlying real value} $r_i^a$ whose distribution is predicted by the calibration network.%
\footnote{That is, $\phat_a(y_i \mid T, Q_i) = \phat_a(r_i \in \mathrm{bin}_{y_i} \mid T, Q_i)$ where $r_i^a$ has a normal (or logistic) distribution whose 2 parameters are predicted from $T$ by the calibration network for $a$, and where the $n$ bins are a partition of $\mathbb{R}$ by $n-1$ learned thresholds that are specific to $a$ or to $(a,i)$.  This gives a discrete distribution over $y_i^a$, which can be used in the log-likelihood objective \labelcref{eq:loglik}. This is a nonlinear, heteroskedastic version of ordered probit (or logit) regression.}
Furthermore, we could then (if desired) evaluate the text $T$ using our best prediction of the underlying $r_i^a$ rather than of the observed $y_i^a$ (e.g., using the expected value as before, if we wish to minimize expected $L_2$ loss).  The intuition is that the reconstructed unquantized $r_i^a$ values contain more information than their quantized versions---and that they might also be more comparable across judges $a$, if their different judgments (e.g., in \cref{fig:judge_scores_dist}) mainly arise from different quantization boundaries.\footnote{However, a trick is needed to ensure that $r_i^a$ values are interpretable and comparable across judges $a$.  The issue is that the method in the previous footnote does not identify the position or scale of $r_i^a$.  (If we adjusted our model to double the predicted means, predicted standard deviations, and thresholds for judge $a$, we would get exactly the same distribution over observables $y_i^a$ and achieve the same log-likelihood.  But $r_i^a$ would now have twice the range and so would count more in a mean over judges $\A$.)  To break this tie, we can augment the log-likelihood objective with a second term (perhaps with infinitesimal weight) that \emph{does} care about position and scale.  Assuming that our ordinal scale $\Y_i$ is numeric, a natural choice for this second term is the \emph{unquantized} log-likelihood: that is, we ask the normal curve to assign a high log-density to the exact value $y_i^a$ and not just a high log-probability to its bin.  This ties $r_i$ to the $y_i$ scale, making it interpretable.}

\paragraph{Comparative Judging.} 
Our maximum-likelihood training principle \labelcref{eq:loglik} can be extended to other question types.  In particular, $Q_i=\mbox{}$``Does $T$ or $T'$ score higher on criterion $i$?\@''
can be interpreted as a comparison of underlying real values as in the preceding paragraph: the human judge is being asked whether $r_i > r'_i$.  The calibration network could predict the probability of a ``yes'' response either directly, or else 
by integrating over a latent distribution $\phat_a(r_i, r_i' \mid T, T', Q_i)$ (perhaps modeling it as $\phat_a(r_i \mid T, Q_i) \cdot \phat_a(r'_i \mid T', Q_i)$ via an independence assumption).\footnote{Unfortunately, any reporting of the predicted $r_i^a$ values (e.g., $\rhat_i^a$) as quality metrics runs into the same non-identifiability problem as in the previous footnote.  We cannot know the position or scale of a judge's $r_i^a$ values if we only observe the results of $>$ comparisons.  A simple fix is to apply an affine transform to each judge's $r_i^a$ values so that on a given reference set of texts, the transformed values have mean 0 and variance 1.  Then report these transformed values.}

\section{\llmeval Questions}
\label{app:questions}

These were the questions in our evaluation rubric.  The human and LLM prompts in which these questions appeared are given in \cref{app:prompt:llm} and \cref{app:prompt:human} respectively. 
When presenting the questions to the LLM (\cref{app:prompt:llm}), boldface was omitted.  When presenting them to human judges on real data (\cref{app:real_data_eval}), boldface was again omitted, and the response choices were not numbered; instead, radio buttons were used (\cref{fig:ui-2}). 

Question instances where the correct answer was ``NA'' were not included in our training dataset $\dataset$ and were not used for evaluation.

\horizbar
\begin{color}{blue}
\noindent \textcolor{black}{$Q_1$--} In terms of naturalness and tone of the \textbf{assistant utterances}, to what degree are they likely to be produced by an \textbf{intelligent human} in a conversation? Disregard whether they are grounded in the search results.

1.	Unlikely.

2.	Somewhat unlikely.

3.	Somewhat likely.

4.	Likely.

\medskip

\noindent \textcolor{black}{$Q_2$--} If the references are provided, to what degree user's questions can be answered or resolved using the references? The assistant’s responses should not impact your response to this question. \textbf{If no references are provided in the conversation, please write “NA” for this question.}  

1.	None of the questions that user has asked could be answered using the reference documents.

2.	Less than half of documents that user has asked could be answered using the reference document.

3.	Half or more than half of the questions that user has asked could be answered using the reference documents.

4.	All the questions the user has asked could be answered with the reference documents.

\medskip

\noindent \textcolor{black}{$Q_3$--}  Independent of what sources are cited in the conversation, to what degree the claims made by the assistant are followed by a citation. \textbf{If no references are provided in the conversation, please write NA.}

1.	None of the claims are followed by a citation.

2.	Less than half of the claims are followed by a citation.

3.	Half, or more than half of the claims are followed by a citation.

4.	All claims are followed by a citation.

\medskip

\noindent \textcolor{black}{$Q_4$--}  What percentage of citations accurately support the claims made in the conversation?  \textbf{If no references are provided in the conversation, please write NA.}

1.	None of the citations accurately support the provided claims.

2.	Less than half of citations accurately support the provided claims.

3.	Half, or more than half of citations accurately support the provided claims.

4.	All citations accurately support the provided claims.

\medskip

\noindent \textcolor{black}{$Q_5$--}  To what degree the cited sources are the best candidates among all the provided sources? \textbf{If no references are provided in the conversation, please write NA.}

1.	For all citations, there is a better source to be cited.

2.	For more than half of the citations, there is a better source to be cited.

3.	 For half or less than half of the citations, there is a better source to be cited.

4.	 The best sources are cited in all cases.

\medskip

\noindent \textcolor{black}{$Q_6$--}  To what degree the content of the assistant utterances is free of redundant elements, such as \textbf{repetition}, \textbf{overspecification}, etc.

1.	The conversation has \textbf{a large number} of redundant elements.

2.	The conversation has \textbf{some} redundant elements.

3.	The conversation has \textbf{a few} redundant elements.

4.	The conversation \textbf{is completely free} of redundant elements.

\medskip

\noindent \textcolor{black}{$Q_7$--}  To what degree the assistant responses are concise?

1.	In all assistant utterances, the responses could have been shorter.

2.	In more than half of the assistant utterances, the responses could have been shorter.

3.	In half, or less than half of the assistant utterances, the responses could have been shorter.

4.	In all assistant utterances, the responses are concise and the utterance length is appropriate.

\medskip

\noindent \textcolor{black}{$Q_8$--}   Do you think the number of exchange turns or back and forth is appropriate given the complexity of the user information need?\textcolor{black}{\footnote{For $Q_8$, the numeric responses unfortunately do not form an ordinal scale.  Response \texttt{``3''} should reasonably be considered closer to \texttt{``1''} than it is to \texttt{``2''}.  Thus, $L_2$ is not an appropriate loss function here.  However, for simplicity we did still use $L_2$ when decoding $\yhat_8^a$ (it motivates \cref{eq:mbr}) and when evaluating the quality of  $\yhat_8^a$ (it motivates RMSE).  This affects only the $Q_8$ line of \cref{tab:per_dim_results}, all of whose metrics would presumably be improved if we fixed the problem by swapping \texttt{``2''} and \texttt{``3''} in both the human data and the LLM data.  All of our other results would be unaffected by this relabeling.}}

1.	No, fewer interactions would be sufficient and would make this conversation more pleasant.

2.	No, more interactions are needed for a better conversation experience.

3.	Yes, the rate of exchanges between the user and the assistant is reasonable.
\end{color}

\horizbar

\begin{color}{blue}
\noindent \textcolor{black}{$Q_0$--}  Imagine you are the user who had this conversation with the assistant. All in all, how you would rate your overall satisfaction while interacting with the assistant? The higher the rating, the better the experience.

1.	1

2.	2

3.	3

4.	4

\end{color}
\horizbar

\section{Evaluation Prompt for LLM}
\label{app:prompt:llm}
In our \llmeval experiments (\cref{sec:exp_setup}), we use the following prompt template to ask the LLM an evaluation question $Q_i$ about a conversational text $T$.

The variable \{conversation\} is the complete dialogue between the user and the assistant, and the variable \{question\} is one of the questions from the evaluation rubric presented in \cref{app:questions}.

The citation-related questions $Q_2$, $Q_3$, $Q_4$, and $Q_5$ are not presented to the LLM if no references are provided in the conversation.  In this case, we simply pretend that the LLM would have correctly answered ``NA,'' which means that the probability vector over the responses 1--4 is $[0, 0, 0, 0]$ (see \cref{fn:unnorm}).

\horizbar
\begin{color}{blue}

\noindent You are given a conversation between a user and an intelligent assistant for an enterprise chat scenario. In some cases, some references and citations are provided to back up the claims made by the intelligent assistant. Your primary job is to evaluate the quality of the conversation based on a criterion. To do so, read the conversation and references, and answer the followed question, by selecting only one of the choices.

Conversation: \textcolor{black}{\{conversation\}}

Question: \textcolor{black}{\{question\}}

Only print '1', '2', '3', or '4'.

\end{color}
\horizbar

\section{Evaluation Prompt and Preliminary Data Quality Questions for Humans}
\label{app:prompt:human}

Below are the instructions we gave to human judges with the main questions in \cref{app:questions}.

The preliminary questions DQQ0--DQQ2 are used only to screen for problems with the generated synthetic dialogues of \cref{sec:data:synthetic} (see \cref{app:data_gen_eval}).  They are not included when the human is judging the real dialogues of \cref{sec:data:real}.  Note that if the answer to DQQ is ``No,'' then the remaining questions are not answered, which is why our synthetic training dataset $\dataset$ had only \num{741} examples rather than \num{750}.

\horizbar
\begin{color}{blue}

You are given a \textbf{conversation} between a user and an intelligent assistant for an enterprise chat scenario. You are also given \textbf{an information need that the user wants to fulfill through the course of the conversation (e.g., a problem the user faces and wants to resolve)}. In some cases some references and citations are provided to back up the claims made by the intelligent assistant. Each assistant utterance can only cite the references listed in the adjacent cell in the table. 

\noindent \textbf{Your primary job is to evaluate the quality of the conversation through a series of criteria that we define later in the document. To evaluate the conversation, you need to answer a questionnaire. Each question captures one evaluation criteria that we care about.} 

\noindent Read about the definition of labels criteria below:

\medskip

\noindent \textbf{Naturalness (both content and form)}: The degree to which the form and content of the conversation is realistic, and likely to happen in real-world. To measure naturalness you should answer below questions:

\medskip

\noindent \textcolor{black}{DQQ0-}
Is this a conversation between a user and an assistant?

1. Yes

2. No (if you select ‘No’, you can skip the rest of the questions)

\medskip

\noindent \textcolor{black}{DQQ1-} To what degree the user tries to fulfill the information need during the course of conversation? 

1. The conversation is not about the user information need at all.

2. The conversation does not exactly address the user information need, but it is somewhat related.

3. The conversation addresses the user information need but it also talks about other topics.

4. The conversation only addresses the user information need.

\medskip

\noindent \textcolor{black}{DQQ2-}  To what degree the form and content of the \textbf{user utterances} are likely to be produced by a \textbf{human} in a conversation?

1.	Unlikely.

2.	Somewhat unlikely.

3.	Somewhat likely.

4.	Likely.

\medskip
\noindent\textcolor{black}{\{$Q_1$\}}
\medskip

\noindent \textbf{Citation quality}: To what degree the claims made by the \textbf{assistant} are backed by reliable sources.  Note that not all the sentences in a conversation require citation; only facts and claims need to be cited. To measure citation quality answer the following questions:

\medskip
\noindent\textcolor{black}{\{$Q_2$\}}

\noindent\textcolor{black}{\{$Q_3$\}}

\noindent\textcolor{black}{\{$Q_4$\}}

\noindent\textcolor{black}{\{$Q_5$\}}
\medskip

\noindent\textbf{Dialogue efficiency}: To what degree the dialogue has been conducted in an cost effective manner. To measure the dialogue efficiency answer the following questions:

\medskip
\noindent\textcolor{black}{\{$Q_6$\}}

\noindent\textcolor{black}{\{$Q_7$\}}

\noindent\textcolor{black}{\{$Q_8$\}}

\medskip

\noindent\textbf{User Satisfaction}: Answer the following question to rate the overall user experience with the assistant.

\medskip
\noindent\textcolor{black}{\{$Q_0$\}}

\end{color}
\horizbar

\section{Synthetic Dialogue Generation}
\label{app:data_gen}

This section describes the 5 approaches that we used in \cref{sec:data:synthetic} to generate a variety of synthetic dialogues.

\paragraph{DS1: LLM-Only Assistant with Simulated User.}
In our baseline, the dialogue system has no access to external documents and can only answer the user from its internal knowledge. In this setting, the assistant cannot provide citations for its claims. 

\paragraph{DS2: Oracle RAG Assistant with Oracle Simulated User.}
In this variant, the prompt includes highly relevant documents: the 5 documents that were most frequently clicked when the given topic appeared as a query in the real logs of \cref{sec:data:log_mining}.
Thus, the assistant is essentially a RAG system with unrealistically good retrieval.  In addition, the simulated user is unrealistically knowledgeable, having full access to the same documents for the initial question and all followup questions.

\paragraph{DS3: RAG Assistant with Oracle Simulated User.}
This variant resembles DS2, except that it uses the 5 documents that are most similar to the topic string according to the BM25 metric. 
We use the ElasticSearch\footnote{\url{https://www.elastic.co/}} implementation of BM25.

\paragraph{DS4: RAG Assistant with Simulated User.}
This variant resembles DS3, but the topic is included in the prompt only when generating simulated user turns, and the 5 documents are included in the prompt only when generating assistant turns.  In addition, the BM25 query is not the topic string but rather the dialogue history (all past utterances); thus, each assistant turn may be prompted using a different set of 5 documents.

\paragraph{DS5: Retrieval-Augmented Dialogue Generation + Query Generation with Simulated User.}
This variant resembles DS4, but the BM25 query is not the dialogue history.  Instead, it is derived from the dialogue history by a separate prompt to the LLM (also shown in Table~\ref{tab:data_gen_prompts}).  This may be required as calling a \emph{query generation tool}.

\bigskip
The prompts used for synthetic dialogue generation (DS1--DS5) are presented in \cref{tab:data_gen_prompts}.

\section{Quality of the Generated Synthetic Dialogues}
\label{app:data_gen_eval}

\begin{table}[t]
    \centering
    \begin{tabular}{lccccccc}\hline
        & \textbf{DS1} & \textbf{DS2} & \textbf{DS3} & \textbf{DS4} & \textbf{DS5} \\\hline
        DQQ1 & \num{3.8 \pm 0.5} & \num{3.6 \pm 0.7} & \num{3.6 \pm 0.8} & \num{3.6 \pm 0.8} & \num{3.5 \pm 0.9} \\
        DQQ2 & \num{3.6 \pm 0.6} & \num{3.4 \pm 0.8} & \num{3.3 \pm 0.9} & \num{3.3 \pm 0.9} & \num{3.3 \pm 0.8} \\\hline
        $Q_1$ & \num{3.4 \pm 0.7} & \num{3.2 \pm 0.9} & \num{3.2 \pm 0.9} & \num{3.1 \pm 0.8} & \num{3.1 \pm 0.9} \\
        $Q_2$ & NA & \num{3.5 \pm 0.8} & \num{3.3 \pm 0.9} & \num{3.4 \pm 0.9} & \num{3.3 \pm 1.0} \\
        $Q_3$ & NA & \num{3.3 \pm 0.9} & \num{3.0 \pm 1.0} & \num{3.2 \pm 1.1} & \num{3.0 \pm 1.3} \\
        $Q_4$ & NA & \num{3.3 \pm 0.8} & \num{3.1 \pm 1.0} & \num{3.1 \pm 1.1} & \num{3.0 \pm 1.3} \\
        $Q_5$ & NA & \num{3.3 \pm 0.9} & \num{3.1 \pm 1.0} & \num{2.9 \pm 1.1} & \num{2.6 \pm 1.3} \\
        $Q_6$ & \num{3.6 \pm 0.7} & \num{3.3 \pm 0.8} & \num{3.2 \pm 0.9} & \num{3.7 \pm 0.6} & \num{3.6 \pm 0.7} \\
        $Q_7$ & \num{3.6 \pm 0.7} & \num{3.1 \pm 0.9} & \num{3.1 \pm 1.0} & \num{3.5 \pm 0.8} & \num{3.6 \pm 0.7} \\
        $Q_8$ & \num{2.7 \pm 0.7} & \num{2.5 \pm 0.9} & \num{2.6 \pm 0.8} & \num{2.5 \pm 0.6} & \num{2.5 \pm 0.6} \\
        $Q_0$ & \num{2.3 \pm 0.7} & \num{3.1 \pm 0.8} & \num{3.0 \pm 0.8} & \num{3.0 \pm 0.9} & \num{2.9 \pm 0.9} \\\hline
    \end{tabular}
    \caption{Mean and standard deviation of human annotations for different sets of synthetic dialogues.  As each column has $n \approx 148$ examples, the standard error of the mean is about $\frac{1}{12}$ of the standard deviation shown.  Thus a 95\% confidence interval on the mean is $\pm\frac{1}{6}$ of the standard deviation, ranging here from \num{\pm 0.1} to \num{\pm 0.2}.}   \label{tab:dialogue_annotation}
\end{table}

\begin{table}[t]
    \centering
    \begin{tabular}{lccccc}\hline
        & \textbf{DS1} & \textbf{DS2} & \textbf{DS3}  \\\hline
        \# conversation & \num{76} & \num{71} & \num{76} \\\hline
        $Q_1$ & \num{3.329 \pm 0.817} & \num{3.197 \pm 0.973} & \num{3.066 \pm 0.964} \\
        $Q_2$ & NA & \num{3.155 \pm 0.816} & \num{2.763 \pm 0.930} \\
        $Q_3$ & NA & \num{2.971 \pm 0.903} & \num{2.631 \pm 0.900} \\
        $Q_4$ & NA & \num{3.112 \pm 0.943} & \num{2.618 \pm 0.959} \\
        $Q_5$ & NA & \num{3.014 \pm 1.013} & \num{2.631 \pm 1.049} \\
        $Q_6$ & \num{3.473 \pm 0.734} & \num{3.436 \pm 0.686} & \num{3.473 \pm 0.678} \\
        $Q_7$ & \num{3.552 \pm 0.768} & \num{3.295 \pm 1.012} & \num{3.500 \pm 0.716} \\
        $Q_8$ & \num{2.803 \pm 0.487} & \num{2.788 \pm 0.501} & \num{2.739 \pm 0.440} \\
        $Q_0$ & \num{2.668 \pm 0.817} & \num{2.915 \pm 0.835} & \num{2.697 \pm 0.707} \\
        \hline
    \end{tabular}
    \caption{Mean and standard deviation of different sets of real human-agent dialogues.  In all cases, $\pm$ \num{0.24} gives a \num{95}\% confidence interval on the mean.}
    \label{tab:real_dialogue_annotation}

\end{table}

As mentioned in \cref{sec:data:synthetic}, each of the systems DS1--DS5 (\cref{app:data_gen}) was used to generate 50 synthetic dialogues, each of which was evaluated by 3 human judges, resulting in $5 \times 50 \times 3 = 750$ completed questionnaires.  The first question we asked (DQQ0) was ``Is this a conversation between a user and an assistant?'' As expected based on the findings presented in \citep{li-etal-2023-autoconv}, the answers to this question were vastly positive: \num{98.8}\% of the dialogues received a positive answer.  

\Cref{tab:dialogue_annotation} shows the mean and standard deviation of the human judgments for the questionnaires that passed the DQQ0 quality check.  The results on DQQ1 and DQQ2 suggest that all systems often simulated the user turns competently, and the results on $Q_1$--$Q_8$ and $Q_0$ suggest that all systems often produced reasonably good assistant responses to these simulated users.  
In fact, the DS2 and DS3 systems obtained an average $\geq$ \num{3.0} over their dialogues for \emph{all} questions (except for $Q_8$, where the response scale is 1--3).  

Of course, the point of our \llmeval experiments is not to generate good dialogues but to determine which dialogues show more satisfactory behavior by the assistant.  These generated dialogues simply provide synthetic training and development data for that task.

\paragraph{Questions on the naturalness of dialogues (DQQ1, DQQ2, $Q_1$).}
\Cref{tab:dialogue_annotation} indicates that system DS1 produces the most natural conversations. This is a non-RAG system that simply asks the LLM to write a plausible dialogue on the given topic. The other four systems perform comparably in terms of generating natural dialogues. DS2 performs slightly better than the rest; this it may be due to the high quality of its references, which can be less noisy and confusing than the other variants.

\paragraph{Questions on citations ($Q_2$, $Q_3$, $Q_4$, $Q_5$).} On citation quality and usage, DS2 achieves the highest average rating, thanks to its ``oracle'' RAG.
Among the methods that perform RAG with various BM25 queries, DS3 and DS4 perform slightly better than DS5, which prompts an LLM to generate the BM25 query.

\paragraph{Questions on conciseness ($Q_6$, $Q_7$, $Q_8$).}
All systems are similar at generating an appropriate number of turns ($Q_8$).  DS2 and DS3 seem to have less concise dialogues ($Q_6$, $Q_7$), perhaps because the simulated user has access to the retrieved documents.

\paragraph{Question on overall satisfaction ($Q_0$).}
The results suggest that the quality of retrieved documents is the most important factor for our judges, with DS1 clearly doing worst and DS2 doing slightly better than the others.

\begin{figure}[!ht]
    \centering
    \begin{subfigure}{\textwidth}
    \includegraphics[width=0.75\textwidth,clip,trim=0cm 1cm 5cm 1cm]{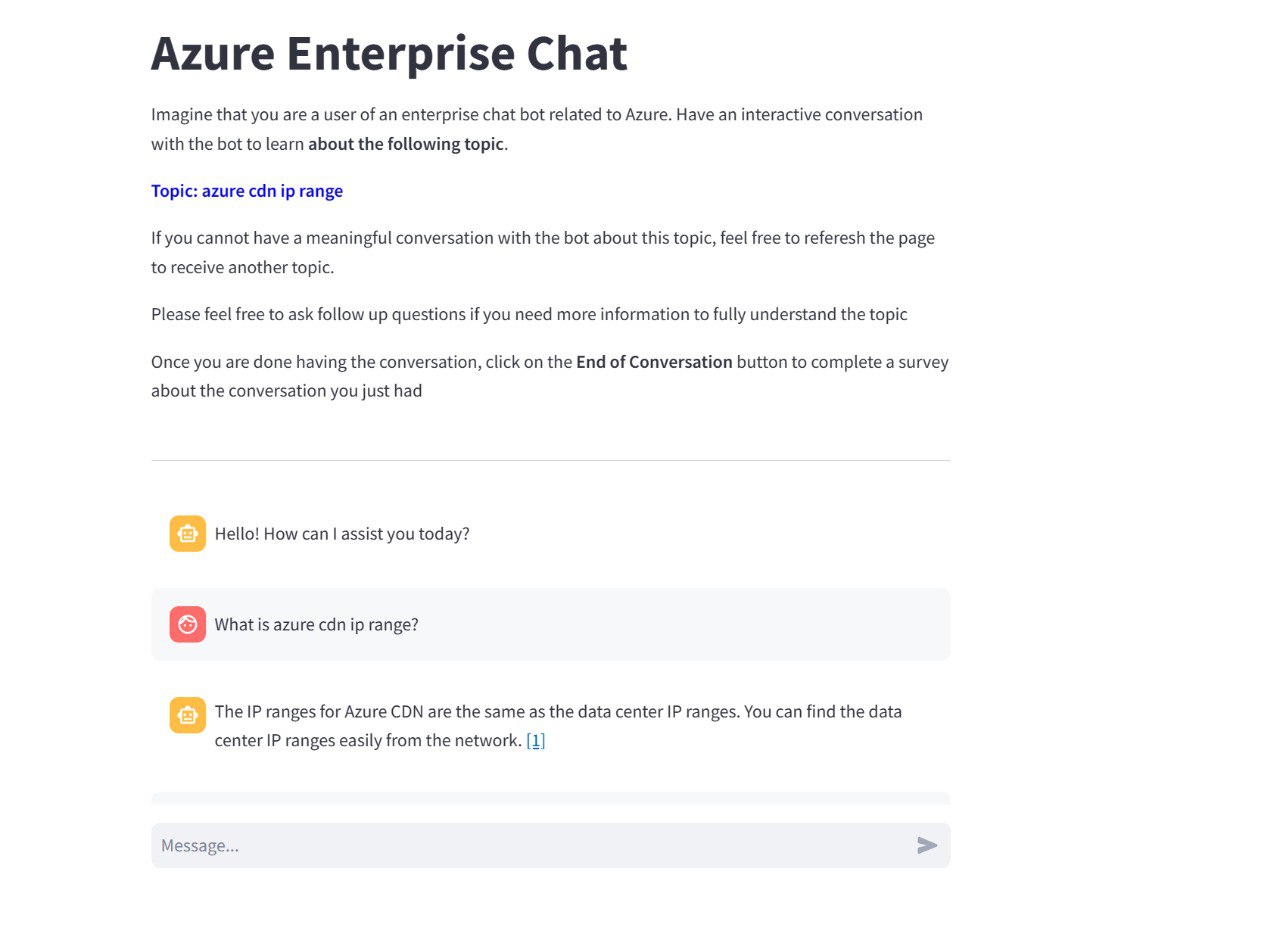}
    \caption{First, we ask the user to have a conversation with the agent about a given topic.}
    \label{fig:ui-1}
    \end{subfigure}
   \medskip
   \begin{subfigure}{\textwidth}
     \includegraphics[width=0.8\textwidth,clip,trim=0cm 1cm 2cm 0cm]{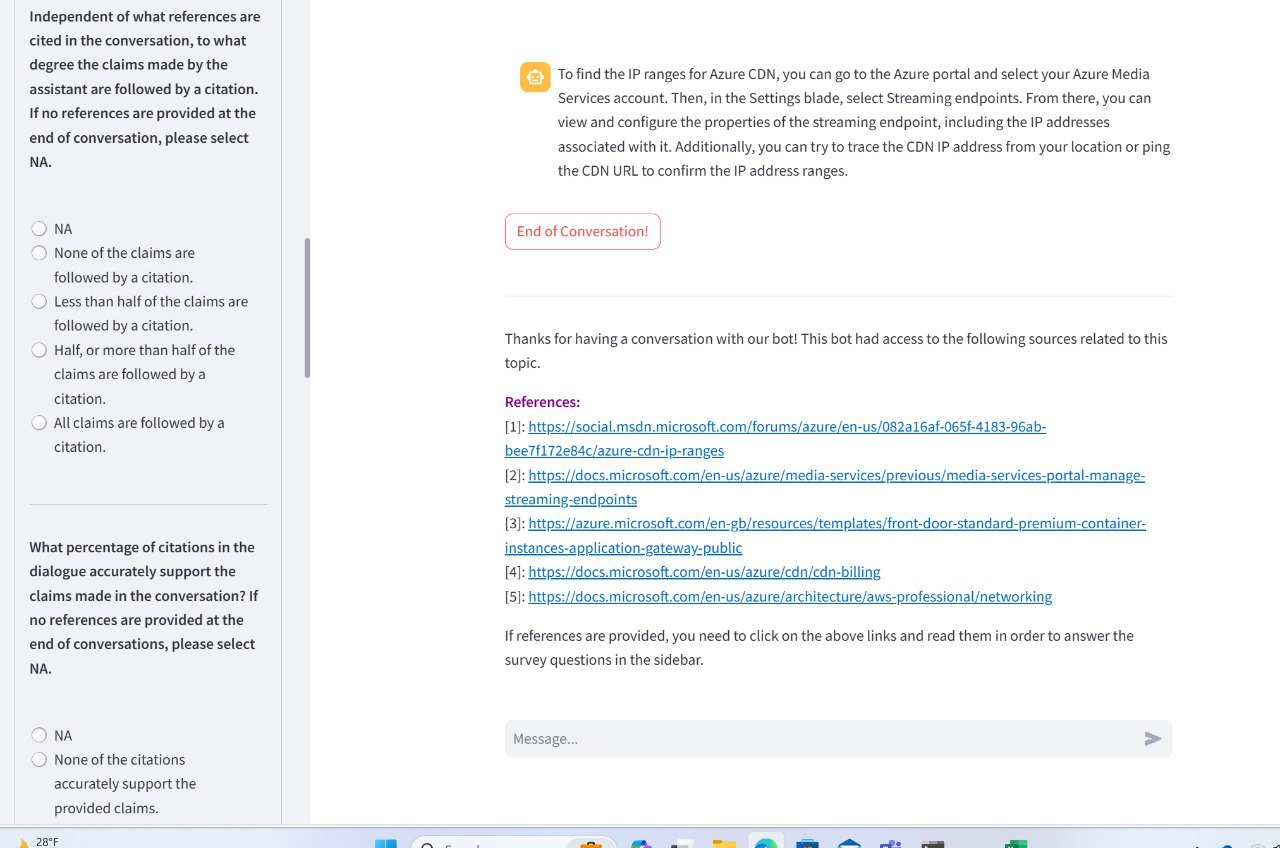}
    \caption{Once the user clicks on 'End of Conversation', we present all the search engine results available to the agent and ask the user to read them. Finally, we ask the user to evaluate their experience with the agent by answering the evaluation rubric questions (\cref{app:questions}).}
    \label{fig:ui-2}
   \end{subfigure}
   \caption{User interface for real dialogue collection and evaluation.}\label{fig:ui}
\end{figure}

\section{The User Interface for Human-Agent Dialogue Collection and Evaluation}
\label{app:interface}

We designed a web interface (\cref{fig:ui}) to enable humans to converse with a dialogue system as users and then evaluate their interactions as judges (\cref{sec:data:real}).
In each session, the website shows the human a random Azure-related topic from the set described in \cref{sec:data:log_mining}, and randomly selects one of the dialogue systems DS1--DS3 for the human to converse with.  The human does not know which system was selected (although DS1 might be distinguishable as it does not give citations).  

This is a standard guided data collection procedure that has been previously used in prior work \citep{info-seeking-conv}.  If the user does not understand the topic, they may refresh the website to get a different topic.  Once the user is finished with their conversation, they click on the `End of Conversation' button and judge the conversation (see \cref{app:questions}).

\section{Evaluating the Collected Human-Agent Dialogues}
\label{app:real_data_eval}

We asked 13 trained judges to use the website for dialogue collection and evaluation. The judge set for the synthetic dialogues presented above includes these 13 judges. We collected a total of 223 conversations, ranging from 14--27 conversations per judge. The judge scores for the three dialogue systems evaluated are summarized in \cref{tab:real_dialogue_annotation}.

\section{How much human judge data is needed to train calibration?}\label{app:learning_curves}

\begin{figure}[t]
    \centering
    \includegraphics[width=0.75\textwidth,clip,trim="2cm 10cm 3cm 10cm"]{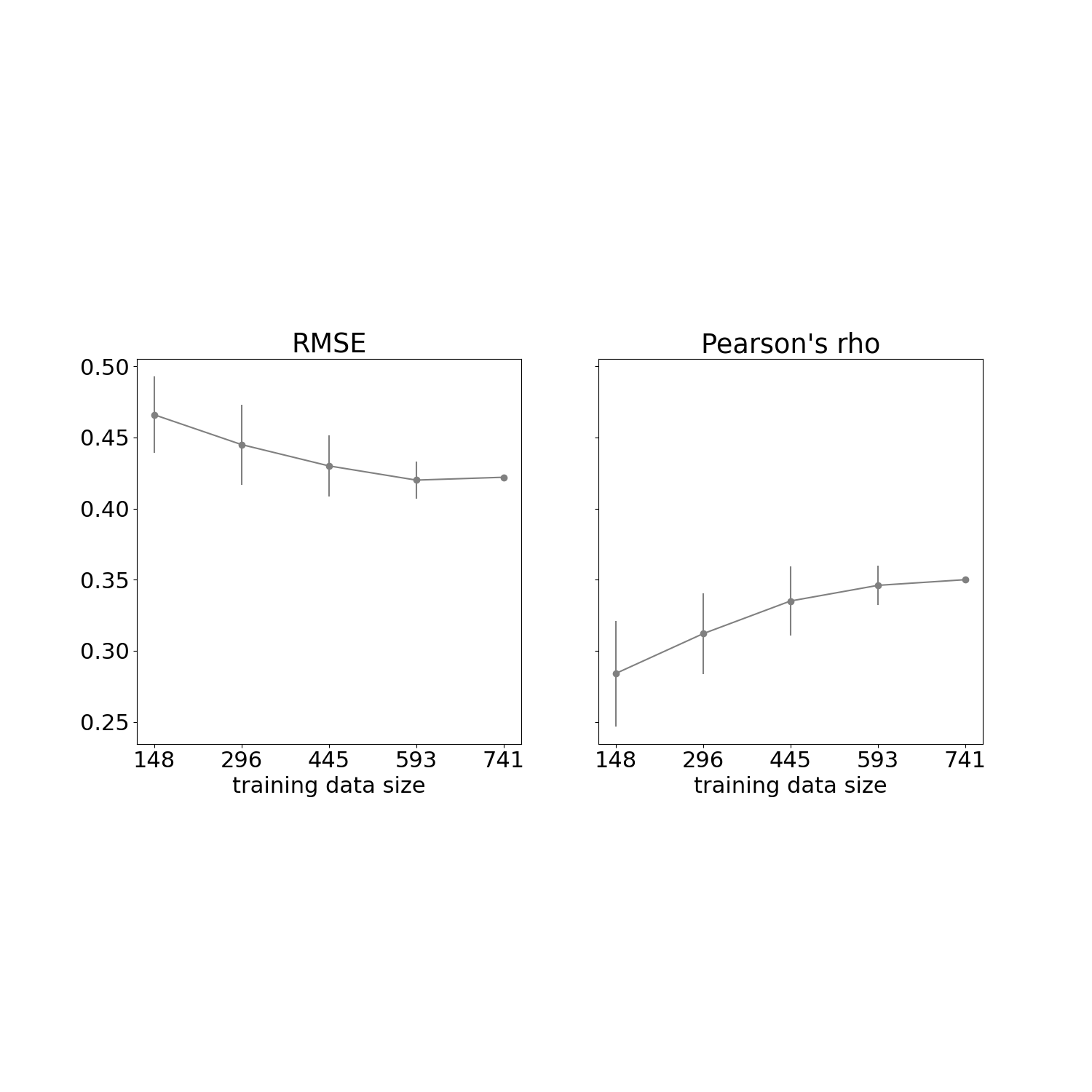}
    \caption{Learning curve for training the personalized calibration network in \llmeval on synthetic conversations and testing on the real conversation data.  The model's performance becomes relatively stable after observing 80\% of the training data.  Note that the LLM itself is not fine-tuned to predict any judge's responses.}
    \label{fig:learning_curve}
\vspace{-.5cm}
\end{figure}

We plot learning curves in \cref{fig:learning_curve}.
To make these plots, we train the model on the synthetic data and test on the real conversation data, but reduce the training portion of the data to a random $x\%$ sample. To reduce the impact of random selection, we repeat this process 50 times and plot the average performance, $\pm$ \num{1} standard deviation.  As expected, the average performance improves and drops in variance as we increase the amount of training data per judge.\footnote{The reduction in variance is partly because the larger training sets are more similar to the population and thus to each other, but also because they overlap more and thus are less independent.}

Performance is reasonably good with even 20\% of our training set, and appears to have essentially converged by the time we reach \num{80}--\num{100}\% of our training set.  (Here \num{100}\% represents \num{741} dialogues, where each judge has evaluated only $\sim$\num{30} dialogues on average.)
Further improvements would, therefore, require more dimensions or more accurate modeling of each dimension, or perhaps training data targeted at fixing the residual errors.

\section{Calibration Plots (Reliability Diagrams)}\label{app:calibration}

\begin{figure}[!ht]
    \centering
    \includegraphics[width=0.23\textwidth]{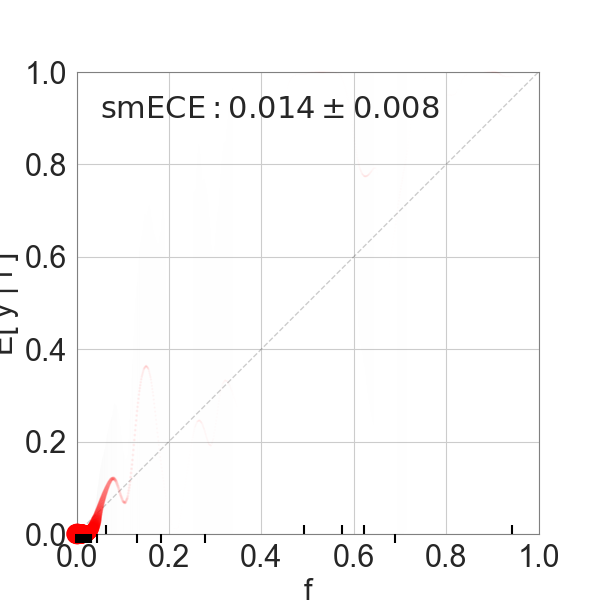}
    \includegraphics[width=0.23\textwidth]{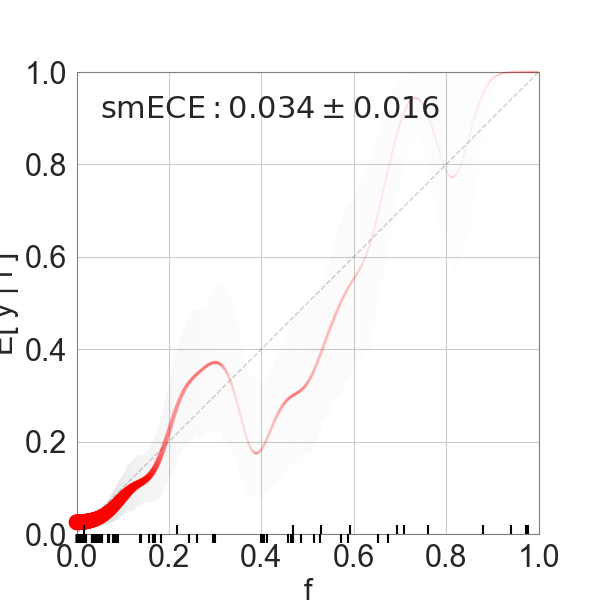}
    \includegraphics[width=0.23\textwidth]{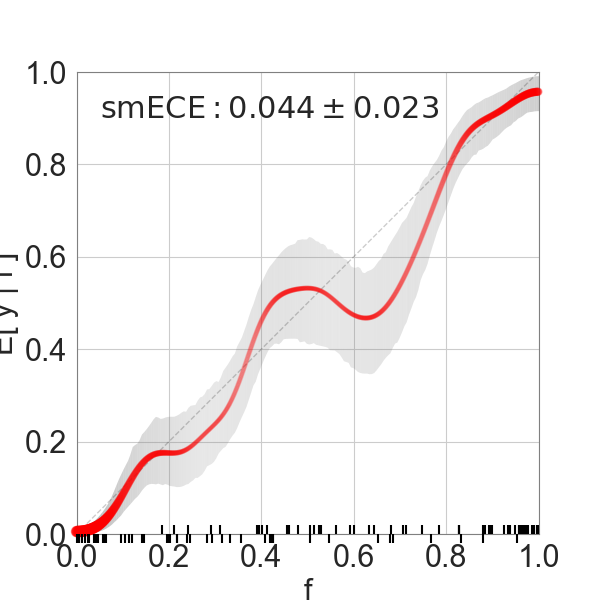}
    \includegraphics[width=0.23\textwidth]{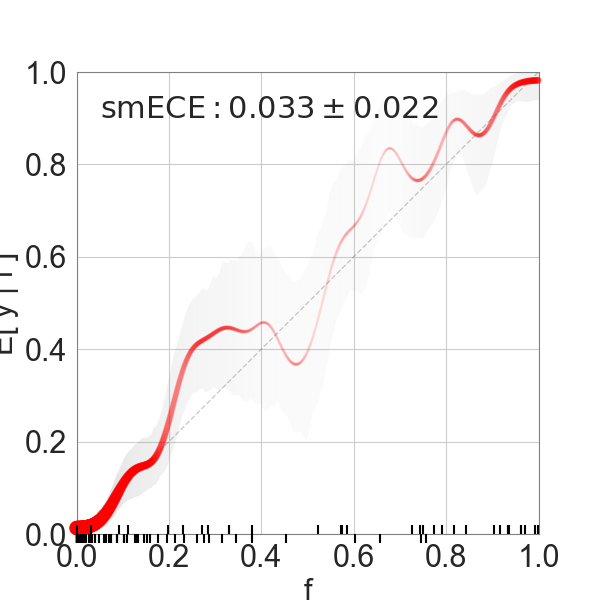}
    \caption{Calibration plots for $Q_0$ on held-out synthetic dialogues, as explained in \cref{sec:analysis,app:calibration}.  These are plots for $y_0 \in \{1,2,3,4\}$ respectively.  They show low calibration error.}\label{fig:calibration}
\end{figure}

Well-trained neural networks
tend to produce well-calibrated probabilities \cite{niculescu2005predicting}, an outcome that is incentivized by
the log-likelihood training and validation objectives.
\Cref{fig:calibration} shows smoothed calibration plots when training and evaluating our system on synthetic dialogues, as explained in \cref{sec:analysis}.  The system is indeed well-calibrated, meaning that the red curves stay close to the diagonal, as measured by smoothed expected calibration error (smECE).

The plots are produced by the \texttt{relplot} package\footnote{\url{https://github.com/apple/ml-calibration}} of \citet{blasiok2023smooth}, using 5-fold cross-validation.  All graphs plot the same examples, namely the tuples $(T,i,a,y_i^a) \in \dataset$, where $T$ is a synthetic dialogue.  However, each graph considers the calibration for a different possible score $y_0$.  Tick marks just above (or below) the horizontal axis are a sample of held-out examples for which the true judgment $y_i^a$ is (or is not) $y_0$.  Their position along the horizontal axis shows their predicted probabilities $\phat_a(y_0 \mid T, Q_0)$ under cross-validation.  Thus, the tick marks above the axis (true score is $y_0$) tend to appear farther to the right ($y_0$ is predicted with high probability).

For each predicted probability $p$, the height of the red curve estimates the actual probability that $y_i^a = y_0$ among held-out examples where $\phat_a(y_0 \mid T, Q_0) \approx p$.  One may think of this visually as the approximate fraction of tick marks that are plotted near $p$ on the horizontal axis that are just above the axis rather than just below.  The thickness of the red curve at $p$ corresponds to the density of tick marks near $p$.  The gray band around the red curve is a 95\% bootstrap confidence interval.  

The smoothed expected calibration error (smECE) is the average absolute distance between the height of the red curve and the diagonal, weighted by the thickness of the curve.  In other words, it estimates the average difference between the predicted and actual probability for a random held-out example.  The smECE number for each graph is printed on the graph with a 95\% confidence interval.

Calibration of $\phat_a$ is important because it means that the predicted probabilities are meaningful.  The system can use them to assess its own uncertainty and make decisions accordingly.  Some applications in our case:
\begin{itemize}
\item \textbf{Text evaluation.} The expected score \labelcref{eq:mbr} will be approximately unbiased: that is, on average over held-out examples, it will match the judge's actual score.  \Cref{tab:main_results} assesses this match directly.  Beyond expected score, various other interesting quantities that we might derive from $\phat_a$ are also approximately unbiased: for example, the probabilities that the score is $\leq 1$, $\leq 2$, and $\leq 3$.
\item \textbf{Text selection.}  At runtime, a dialogue system might generate several candidate responses, and then choose the one with maximum expected reward.  If the reward has the form $\sum_{i,a} f_{i,a}(y_i^a)$, for any set of reward functions $f_{i,a}$, then its expectation under $\phat$ will be unbiased.  
\item \textbf{Dynamic feature selection.} 
\llmeval can be sped up at test time by asking fewer questions of the LLM.   As briefly mentioned in \cref{sec:future,app:extensions}, $\phat_a$ can be used to greedily choose the question with the greatest information gain---that is, whose answer is predicted to most reduce the variance of the evaluation $\yhat_0$ or most reduce the entropy of a text selection decision.
\item \textbf{Distillation.}
\llmeval can be used to stochastically label a large dataset of naturally occurring texts according to $\phat_a$ (``multiple imputation'').  A faster scoring function can then be trained to have low loss on this dataset.
\item \textbf{Rubric improvement.} $\phat_a$ can be used to identify difficult texts where \llmeval is unsure what judge $a$ would say, or controversial texts where \llmeval predicts that two judges will disagree more than usual.  This can be used to improve the LLM questions or the human questions, respectively.
\end{itemize}

As a caveat, the plots in \cref{fig:calibration} measure calibration only for the dataset as a whole.  One could create calibration plots for subsets of the data to verify that the model remains calibrated within each user category, judge, or dialogue topic that is sufficiently represented in training data---as one would expect with maximum-likelihood training---rather than overestimating probabilities in some categories and underestimating them in others.  The correlation coefficients in \cref{fig:calibration} do show that the predicted scores provide a reasonable ranking of examples.

\begin{table}[t]
    \centering
    \begin{tabular}{>{\raggedright\arraybackslash}p{1.8cm}p{12.5cm}}\hline
         & \textbf{Prompt} \\\hline
        \textbf{DS1} &     \begin{color}{blue}
        A user wants to know about ``\textcolor{black}{\{topic\}}''. Write a conversation between a user and a helpful assistant about user's need.
        \end{color}\\\hline

        \textbf{DS2 \& DS3} & \begin{color}{blue}
        A user wants to know about ``\textcolor{black}{\{topic\}}''. Write a conversation between the user and a helpful assistant about user's need. The assistant should provide factual responses using the following Sources. Cite Sources as needed, like [3] or [2].

        Sources:
        \begin{enumerate}
            \item[[  1]] \textcolor{black}{\{Reference 1\}}
            \item[[  2]] \textcolor{black}{\{Reference 2\}}
            \item[    ] $\cdots$
        \end{enumerate}
        \end{color} \\\hline

        \textbf{DS4 \& DS5 (Init)} & \begin{color}{blue}
        Imagine a user wants to know about ``\textcolor{black}{\{topic\}}'' by talking to an intelligent assistant. What would be the first question that the user asks? Generate the output in this format: ``User: utterance''.
        \end{color}\\\hdashline

        \textbf{DS4 \& DS5 (Assistant)} & \begin{color}{blue}
        Imagine a user is interacting with an intelligent assistant to solve their problem. The assistant should provide factual responses using the following sources, or if that is not possible, asks useful questions to get a better understanding of the user need. Cite Sources as needed, like [3] or [2].

        Sources:
        \begin{enumerate}
            \item[[  1]] \textcolor{black}{\{Reference 1\}}
            \item[[  2]] \textcolor{black}{\{Reference 2\}}
            \item[    ] $\cdots$
        \end{enumerate}

        Complete the following dialogue with only one utterance. If there is no need for a response, only generate ``END OF CONVERSATION!''
        \medskip

        Assistant: How can I help you?

        $\cdots$

        User: \textcolor{black}{\{Last Generated User Utterance\}}

        Assistant:
        \end{color}\\\hdashline

        \textbf{DS4 \& DS5 (User)} & \begin{color}{blue}
        Imagine a user is interacting with an intelligent assistant to solve their problem about ``\textcolor{black}{\{topic\}}''. Complete the following dialogue with only one utterance. If there is no need for a response, only generate "END OF CONVERSATION!"

        \medskip

        Assistant: How can I help you?

        $\cdots$

        Assistant: \textcolor{black}{\{Last Generated Assistant Utterance\}}

        User:
        \end{color}\\ \hdashline 

        \textbf{DS5 (QGen)} & \begin{color}{blue}
        Assume that you plan to answer the user's question in the following conversation:

        \medskip

        Assistant: How can I help you?

        $\cdots$

        User: \textcolor{black}{\{Last Generated User Utterance\}}

        What query will you submit to a search engine to find the answer? Only generate the query.
        \end{color}\\ \hline
    \end{tabular}
    \caption{Prompts used for synthetic data generation using \texttt{gpt-3.5-turbo-16k}.}
    \label{tab:data_gen_prompts}
\end{table}

\end{document}